\documentclass[10pt,twocolumn]{article}

\usepackage{cvpr}
\usepackage{times}
\usepackage{epsfig}
\usepackage{graphicx}
\usepackage{amsmath}
\usepackage{amssymb}
\usepackage{authblk}
\usepackage{tablefootnote}
\usepackage{multirow}
\usepackage{hhline}
\usepackage[pagebackref=true,breaklinks=true,colorlinks,bookmarks=false]{hyperref}

\cvprfinalcopy % *** Uncomment this line for the final submission

\def\cvprPaperID{330} % *** Enter the CVPR Paper ID here

\newcommand{\netshort}{SINet}

\newcommand{\Myhat}[1]{\expandafter\hat#1}

\ifcvprfinal\fi
\begin{document}

%%%%%%%%% TITLE
\title{Attend and Interact: Higher-Order Object Interactions for Video Understanding}
% \title{Attend and Interact: Higher-order Object Interactions for Fine-grained Video Understanding}
% \title{Propose and Disclose: Learning Object Relationships for Fine-grained Video Understanding}

\newcommand*\samethanks[1][\value{footnote}]{\footnotemark[#1]}

\makeatletter
\renewcommand\AB@affilsepx{, \protect\Affilfont}
\makeatother

\author[1]{Chih-Yao Ma\thanks{Work performed as a NEC Labs intern}}
\author[2]{Asim Kadav}
\author[2]{Iain Melvin}
\author[3]{Zsolt Kira}
\author[1]{Ghassan AlRegib}
\author[2]{Hans Peter Graf}
\affil[1]{\small Georgia Institute of Technology}
\affil[2]{\small NEC Laboratories America}
\affil[3]{\small Georgia Tech Research Institute}
% \affil[ ]{\small\textit {\{cyma, zkira, alregib\}@gatech.edu}}
% \affil[ ]{\small\textit {\{asim, iain, hpg\}@nec-labs.com}}

\maketitle
\begin{abstract}
    Human actions often involve complex interactions across several inter-related objects in the scene. 
    However, existing approaches to fine-grained video understanding or visual relationship detection often rely on single object representation or pairwise object relationships.
    Furthermore, learning interactions across multiple objects in hundreds of frames for video is computationally infeasible and performance may suffer since a large combinatorial space has to be modeled.
    In this paper, we propose to efficiently learn higher-order interactions between arbitrary subgroups of objects for fine-grained video understanding. 
    We demonstrate that modeling object interactions significantly improves accuracy for both action recognition and video captioning, while saving more than 3-times the computation over traditional pairwise relationships.
    The proposed method is validated on two large-scale datasets: Kinetics and ActivityNet Captions. 
    Our \netshort\ and \netshort-Caption\ achieve state-of-the-art performances on both datasets even though the videos are sampled at a maximum of 1 FPS. 
    To the best of our knowledge, this is the first work modeling object interactions on open domain large-scale video datasets, 
    and we additionally model higher-order object interactions which improves the performance with low computational costs. 
\end{abstract}

\section{Introduction}

\begin{figure}[t]
    \centering
    \includegraphics[width=0.48\textwidth]{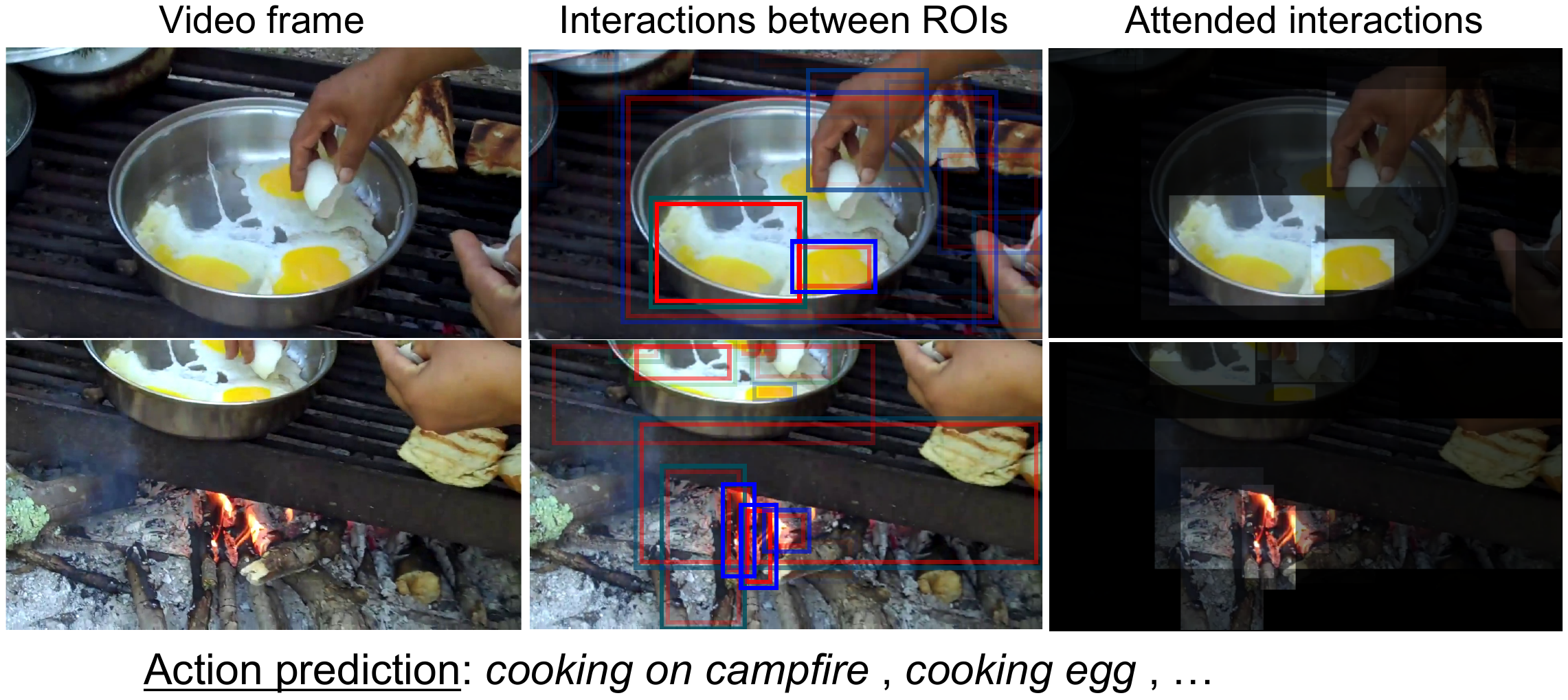}
    \caption{
    \textit{Higher-order object interactions} are progressively detected based on selected inter-relationships.
    % \textit{Top}: consecutive video frames sampled at 1 FPS.
    % \textit{Middle}: video frames with detected relationship regions of interest (ROIs). 
    ROIs with the same color (weighted \textcolor{red}{r}, \textcolor{green}{g}, \textcolor{blue}{b}) indicating there exist inter-object relationships, e.g. eggs in the same bowl, hand breaks egg, and bowl on top of campfire (interaction within the same color).
    Groups of inter-relationships then jointly model higher-order object interaction of the scene (interaction between different colors). 
    \textit{Right}: ROIs are highlighted with their attention weights for higher-order interactions.
    The model further reasons about the interactions through time and predicts \textit{cooking on campfire} and \textit{cooking egg}.
    Images are generated from \netshort\ (best viewed in color).
    }
    \label{fig:concept}
    \vspace{-0.1in}
\end{figure}
Video understanding tasks such as activity recognition and caption generation are crucial for various applications in surveillance, video retrieval, human behavior understanding, etc. 
Recently, datasets for video understanding such as Charades~\cite{sigurdsson2016hollywood}, Kinetics~\cite{kay2017kinetics}, and ActivityNet Captions~\cite{krishna2017dense} contain diverse real-world examples and represent complex human and object interactions that can be difficult to model with state-of-the-art video understanding methods~\cite{sigurdsson2016hollywood}. 
Consider the example in Figure~\ref{fig:concept}. 
To accurately predict \textit{cooking on campfire} and \textit{cooking egg} among other similar action classes requires understanding of fine-grained object relationships and interactions. For example, a hand breaks an egg, eggs are in a bowl, the bowl is on top of the campfire, campfire is a fire built with wood at a camp, etc. 
Although recent state-of-the-art approaches for action recognition have demonstrated significant improvements over datasets such as UCF101~\cite{soomro2012ucf101}, HMDB51~\cite{kuehne2011hmdb}, Sports-1M~\cite{karpathy2014large}, THUMOS~\cite{idrees2017thumos}, ActivityNet~\cite{caba2015activitynet}, and YouTube-8M~\cite{abu2016youtube},
they often focus on representing the overall visual scene (coarse-grained) as sequence of inputs that are combined with temporal pooling, e.g. CRF, LSTM, 1D Convolution, attention, and NetVLAD~\cite{bian2017revisiting,ma2017ts,miech2017learnable,sigurdsson2016asynchronous}, or use 3D Convolution for the whole video sequence~\cite{carreira2017quo, Qiu_2017_ICCV, tran2014c3d}.
These approaches ignore the fine-grained details of the scene and do not infer interactions between various objects in the video. 
On the other hand, in video captioning tasks, although prior approaches use spatial or temporal attention to selectively attend to fine-grained visual content in both space and time, they too do not model object interactions.

% However, most of these datasets focus on sports related actions, e.g. playing tennis, basketball, swimming, etc, which can be distinguished by relying on background or overall scene representation. 

\begin{figure}[t]
    \centering
    \includegraphics[width=0.40\textwidth]{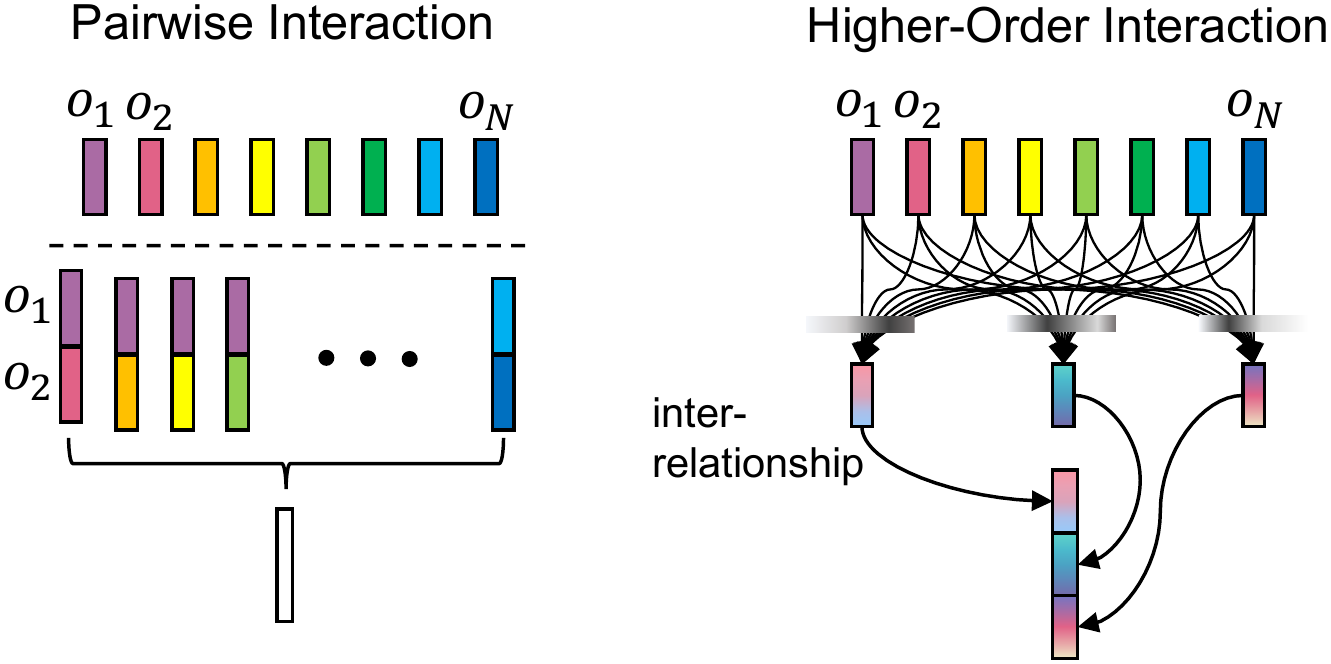}
    \caption{
    Typically, object interaction methods focus on pairwise interactions (left). We efficiently model the \textit{higher-order interactions} between arbitrary subgroups of objects for video understanding, in which the inter-object relationships in one group are detected and objects with significant relationships (i.e. those that serve to improve action recognition or captioning in the end) are attentively selected (right). 
    The higher-order interaction between groups of selected object relationships are then modeled after concatenation. 
    }
    \label{fig:higher-order-interaction}
    \vspace{-0.1in}
\end{figure}

Prior work in understanding visual relationships in the image domain has recently emerged as a prominent research problem, e.g. 
% which has recently developed in image domain, e.g. 
% image captioning~\cite{johnson2016densecap,anderson2017bottom}, visual question answering~\cite{santoro2017simple}, 
scene graph generation~\cite{liang2017deep,xu2017scene} and visual relationship detection~\cite{chao2017learning,dai2017detecting,gkioxari2017detecting,hu2016modeling,zhang2017visual,zhang2017relationship}. 
However, it is unclear how these techniques can be adapted to open-domain video tasks, given that the video is intrinsically more complicated in terms of temporal reasoning and computational demands. 
More importantly, a video may consist of a large number of objects over time. 
Prior approaches on visual relationship detection typically model the full pairwise (or triplet) relationships. 
While this may be realized for images, videos often contain hundreds or thousands of frames.
Learning relationships across multiple objects alongside the temporal information is computationally infeasible on modern GPUs, and performance may suffer due to the fact that a finite-capacity neural network is used to model a large combinatorial space. 
Furthermore, prior work in both image and video domains~\cite{ni2014multiple, ni2016progressively} often focus on pairwise relationships or interactions, where interactions over groups of interrelated objects---\textit{higher-order interactions}---are not explored, as shown in Figure~\ref{fig:higher-order-interaction}. 

Toward this end, we present a generic recurrent module for fine-grained video understanding, which dynamically discovers higher-order object interactions via 
an efficient dot-product attention mechanism combined with temporal reasoning. 
% The proposed model first progressively discover groups of objects based on their inter-relationships, overall image context representation, and past object interactions. 
% The groups of selected objects are then combined to learn their interactions at current video frame. 
Our work is applicable to various open domain video understanding problems.
In this paper, we validate our method on two video understanding tasks with new challenging datasets: action recognition on Kinetics~\cite{kay2017kinetics} and video captioning on ActivityNet Captions~\cite{krishna2017dense} (with ground truth temporal proposals).
% datasets, Kinetics~\cite{kay2017kinetics} and ActivityNet Captions~\cite{krishna2017dense} for action recognition and video captioning tasks.
By combining both coarse- and fine-grained information, our \textbf{\netshort} (Spatiotemporal Interaction Network) for action recognition and \textbf{\netshort-Caption} for video captioning achieve state-of-the-art performance on both tasks while using RGB video frames sampled at only maximum 1 FPS.
% Specifically, we achieved 74.1\% top-1 and 91.5\% top-5 prediction accuracy on Kinetics, and 21.04 ROUGE-L, 9.52 METEOR, and 41.82 CIDEr-D on ActivityNet Captions. 
To the best of our knowledge, this is the first work of modeling object interactions on open domain large-scale video datasets, and we also show that modeling higher-order object interactions can further improve the performance at low computational costs. 
% Code will be made available. 
% Our contributions are summarized as follows:
% \begin{itemize}
%   \item We extend Scale Dot-Product Attention~\cite{vaswani2017attention} to learn coarse-grained video representations.
% %   \item We detect fine-grained higher-order object relationships via efficient dot-product attention using object interrelationships, overall visual scene representation, and object interactions at previous time.
%   \item We propose a generic recurrent Higher-Order Interaction module to progressively detect higher-order object relationships for fine-grained video understanding.
%   \item By combining coarse- and fine-grained information, we achieve state-of-the-art results on both action recognition and video captioning tasks using RGB videos sampled at maximum 1 FPS.
% %   \item To our best knowledge, this is the first work modeling higher-order object interactions on open domain large-scale video datasets.
% \end{itemize}

\section{Related work}
We discuss existing work on video understanding based on action recognition and video captioning as well as related work on detecting visual relationships in images and videos.

\textbf{Action recognition:} 
% Early work on action recognition and video classification involved extracting video features and combining them into a fixed sized video description. A bag of features is extracted from the words over the duration of the video from different spatio-temporal positions and a SVM classifier is used to predict the action~\cite{liu2009recognizing, wang2009evaluation}. 
Recent work on action recognition using deep learning involves learning compact (coarse) representations over time and use pooling or other aggregation methods to combine information from each video frame, or even across different modalities~\cite{feichtenhofer2016convolutional, girdhar2017actionvlad, miech2017learnable, sigurdsson2016asynchronous, simonyan2014two}. The representations are commonly obtained directly from forward passing a single video frame or a short video snippet to a 2D ConvNet or 3D ConvNet~\cite{carreira2017quo, Qiu_2017_ICCV, tran2014c3d}. 
Another branch of work uses Region Proposal Networks (RPNs) to jointly train action detection models ~\cite{gkioxari2015contextual,li2016human, peng2016multi}. 
These methods use an RPN to extract object features (ROIs), but they do not model or learn interactions between objects in the scene. Distinct from these models, we explore human action recognition task using coarse-grained context information and fine-grained higher-order object interactions.
% in an efficient manner using dot-product attention.
Note that we focus on modeling object interactions for understanding video in a fine-grained manner and we consider other modalities, e.g. optical flow and audio information, to be complementary to our method. 

\textbf{Video captioning:}
Similar to other video tasks using deep learning, initial work on video captioning learn compact representations combined over time. This single representation is then used as input to a decoder, e.g. LSTM, at the beginning or at each word generation to generate a caption for the target video~\cite{pan2016jointly, venugopalan2015sequence, venugopalan2014translating}. 
Other work additionally uses spatial and temporal attention mechanisms to selectively focus on visual content in different space and time during caption generation~\cite{ramanishka2017top, song2017hierarchical, yao2015describing, yu2016video, zanfir2016spatio}. 
Similar to using spatial attention during caption generation, another line of work has additionally incorporated semantic attributes~\cite{gan2017semantic, pan2017video, shen2017weakly, yu2017end}. 
However, these semantic or attribute detection methods, with or without attention mechanisms, do not consider object relationships and interactions, i.e. they treat the detected attributes as a bag of words. 
Our work, \netshort-Caption\ uses higher-order object relationships and their interactions as visual cues for caption generation. 

\textbf{Interactions/Relationships in images:}
%Previous works have used relationship between objects to improve object recognition performance~\cite{galleguillos2008object, gould2008multi}. This contrasts with our goals of learning relationships and improving video understanding using existing object detection methods. %
%Pre-deep learning relationships work that are highly-cited: ~\cite{maji2011action, rohrbach2013translating, yao2010modeling}
Recent advances in detecting visual relationships in images use separate branches in a ConvNet to explicitly model objects, humans, and their interactions~\cite{chao2017learning, gkioxari2017detecting}.
Visual relationships can also be realized by constructing a scene graph which uses a structured representation for describing object relationships and their attributes~\cite{johnson2015image, li2017scene, liang2017deep, xu2017scene}.
Other work on detecting visual relationships explore relationships by pairing different objects in the scene~\cite{dai2017detecting, hu2016modeling, santoro2017simple, zhang2017visual}. 
While these models can successfully detect visual relationships for images, a scene with many objects may have only a few individual interacting objects. 
It would be inefficient to detect all relationships across all individual object pairs~\cite{zhang2017relationship}, making these methods intractable for the video domain. 

\textbf{Interactions/Relationships in videos:}
Compared to the image domain, there is limited work in exploring relationships for video understanding.
Ni et al.~\cite{ni2014multiple} use a probabilistic graphical model to track interactions, but their model 
% can only handle the binary contextual relationship between objects.
% which considered mutual geometrical relationship
% This method 
is insufficient to model interactions involving multiple objects.
To overcome this issue, Ni et al.~\cite{ni2016progressively} propose using a set of LSTM nodes to incrementally refine the object detections. 
In contrast, Lea et al.~\cite{lea2016segmental} propose to decompose the input image into several spatial units in a feature map, which then captures the object locations, states, and their relationships using shared ConvNets. 
However, due to lack of appropriate datasets, existing work focuses on indoor or cooking settings where the human subject along with the objects being manipulated are at the center of the image.
Also, these methods only handle pairwise relationships between objects. However, human actions can be complex and often involve higher-order object interactions. 
Therefore, we propose to attentively model object inter-relationships and discover the higher-order interactions on large-scale and open domain videos for fine-grained understanding. 

\begin{figure}[t]
    \centering
    \includegraphics[width=0.48\textwidth]{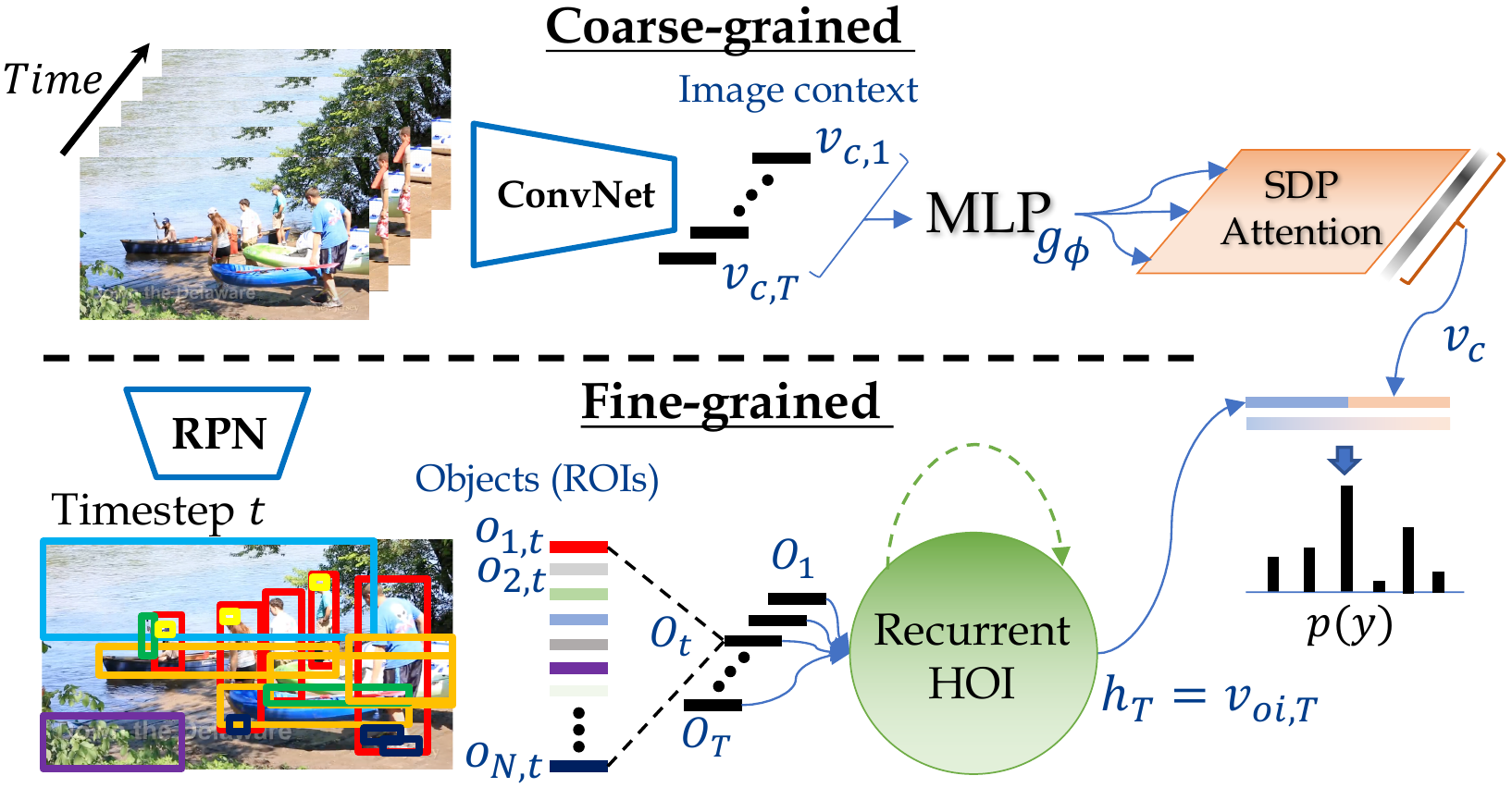}
    \caption{
    Overview of the \netshort\ for action recognition.
    \textbf{Coarse-grained:} each video frame is encoded into a feature vector $v_{c,t}$.
    The sequence of vectors are then pooled via temporal SDP-Attention into single vector representation $v_{c}$.
    \textbf{Fine-grained:} Each object (ROI) obtained from RPN is encoded in a feature vector $o_{n,t}$.
    We detect the higher-order object interaction using the proposed generic recurrent Higher-Order Interaction (HOI) module. 
    Finally, coarse-grained (image context) and fine-grained (higher-order object interactions) information are combined to perform action prediction.
    }
    \label{fig:model-simplified}
    \vspace{-0.2in}
\end{figure}

\section{Model}
% While video understanding has advanced in both action recognition and video captioning, 
Despite the recent successes in video understanding, 
there has been limited progress in understanding relationships and interactions that occur in videos in a fine-grained manner. 
To do so, methods must not only understand the high-level video representations but also be able to explicitly model the relationships and interactions between objects in the scene. 
% Indeed, several approaches have begun to incorporate object-to-object interactions in the image domain~\cite{zhang2017relationship,dai2017detecting,hu2016modeling,chao2017learning,gkioxari2017detecting}.
% However, such methods are infeasible in videos since the set of object-object or object triplet interactions is infeasible to fully represent.
% Furthermore, while detecting object relationship frame by frame is possible, the temporal reasoning of object interactions can not be exploited. 
% Toward this end, we propose to dynamically attend and select objects (ROIs) from a spatial-temporal feature space for discovering higher-order object interactions. 
Toward this end, we propose to exploit both overall image context (coarse) and higher-order object interactions (fine) in the spatiotemporal domain for general video understanding tasks. 
% The \textbf{\netshort} use SDP-Attention to learn coarse-grained context representation and discovers fine-grained higher-order object interactions using image representation and prior object interactions in previous timesteps. 

In the following section, we first describe the \textbf{\netshort} on action recognition followed by extending it to \textbf{\netshort-Caption} for the video captioning task.

\subsection{Action Recognition Model}
\label{sec:model-action}
% \subsubsection{ConvNet + LSTM}
% The ConvNet has been widely used to extract feature representations for images and videos. Typically, in the video domain, this can be easily achieved by, first, using ConvNet to extract feature vectors for each of the video frame independently. Second, the feature vector representation is then used as sequential inputs for the RNNs. One of the commonly used RNNs for video domain is LSTM, which can encode state and model temporal long-term dependencies. The output of the LSTM will then pass to fully connected and softmax layer for action prediction. 

% To understand a human action, it may be sufficient to just rely on the overall visual representation of the scene in some cases. 
% However, for the fine-grained action recognition task, methods must explicitly model the relationships and interactions between objects in the scene. 
% % Note that the \textit{object} denotes here can also imply human and background in the scene.
% Our goal in achieving fine-grained action recognition is to understand both overall image context (coarse) and higher-order object interactions (fine) in the spatiotemporal domain, as illustrated in Figure~\ref{fig:model-simplified}.

\subsubsection{Coarse-grained image context}
% A human action is both defined by the contextual information of the scene and the fine-grained information provided by object interactions in the scene. 
As recent studies have shown, using LSTM to aggregate a sequence of image representations often results in limited performance since image representations can be similar to each other and thus lack temporal variances~\cite{abu2016youtube,kay2017kinetics,ma2017ts}. 
% We suspect this is mainly because the feature vectors obtained from a fine-tuned network for video frames have similar representations since video frames can be quite similar, and the trained LSTM failed to identify key frame in video sequence. 
As shown in Figure~\ref{fig:model-simplified} (top), we thus begin by attending to key image-level representations to summarize the whole video sequence via the Scale Dot-Product Attention (SDP-Attention)~\cite{vaswani2017attention}:
\begin{equation}
\alpha_{c} = softmax(\frac{{X_{c}}^\top {X_{c}}}{\sqrt{d_\phi}}), \quad {X_{c}} = g_\phi (V_{c})
% \alpha_{c} = softmax(\frac{g_\phi (v_{c, t}) \otimes {g_\phi (v_{c, t})}}{\sqrt{d_\phi}}) g_\phi (v_{c, t})
% \alpha_{c} = softmax(\frac{g_\phi (v_{c, t}) {g_\phi (v_{c, t})}^T}{\sqrt{m}}) g_\phi (v_{c, t})
\end{equation}

\begin{equation}
% v_{c}=\sum_{t}^{} \alpha_{c_t}v_{c, t}
v_{c} = \overline{\alpha_c\ {X_{c}}^\top}
\end{equation}
% where $Q, K, V$, in our case, are a sequence image contextual representation for a video, 
% where $W_q \in \mathbb{R}^{n \times L}$ is learned weights, % $L$ is video length, 
where $V_c$ is a set of image features: $V_c = \big\{v_{c, 1}, v_{c, 2}, ..., v_{c, T}\big\}, v_{c, t} \in \mathbb{R}^m$ is the image feature representation encoded via a ConvNet at time $t$, and $t$ ranges from $\big\{1, 2, ..., T\big\}$ for a given video length. 
$g_\phi$ is a Multi-Layer Perceptron (MLP) with parameter $\phi$, 
$d_\phi$ is the dimension of last fully-connected (FC) layer of $g_{\phi}$,
${X_{c}} \in \mathbb{R}^{d_\phi \times T}$ is the projected image feature matrix,
$\sqrt{d_\phi}$ is a scaling factor, 
% and $T$ is the total number of video frame.
and $\alpha_{c} \in \mathbb{R}^{T \times T }$ is an attention weight applied to the (projected) sequence of image representations $V_{c}$.
% Note that the SDP-Attention described above can be further extended to Multi-Head Attention as shown in \cite{vaswani2017attention}, but this is beyond the scope of this work. 
The weighted image representations are then mean-pooled to form video representation $v_c$.

\subsubsection{Fine-grained higher-order object interactions}
\label{sec:obj-interaction}
Traditional pairwise object interactions only consider how each object interacts with another object. 
% As illustrated in Figure~\ref{fig:higher-order-interaction}, we instead explore higher-order interactions. 
% First, we consider an individual object's interrelationships with all other objects. 
% For instance, as shown in Figure~\ref{fig:concept}, the egg has relationships with another egg and with the metal bowl, and wood on the ground has relationships with another wood and with the fire. 
% Second, we further consider interaction between {\it groups} of objects, where each of the groups has interrelationships between their own objects - \textit{higher-order interactions}.
% For example, a hand interacts with the egg in the bowl, or the egg in the bowl interacts with wood with fire. 
We instead model inter-relationships between arbitrary subgroups of objects, the members of which are determined by a learned attention mechanism, as illustrated in Figure~\ref{fig:higher-order-interaction}. 
Note that this covers pair-wise or triplet object relationships as a special case, in which the learned attention only focus on one single object.

\textbf{Problem statement:}
We define \textit{objects} to be a certain region in the scene that might be used to determine the visual relationships and interactions. 
% Our objective is to efficiently detect higher-order interactions from these rich but unordered object representation sets across time. 
Each object representation can be directly obtained from an RPN and further encoded into an object feature. 
Note that we do not encode object class information from the detector into the feature representation since there exists a cross-domain problem, and we may miss some objects that are not detected by the pre-trained object detector. 
Also, we do not know the corresponding objects across time since linking objects through time can be computationally expensive for long videos.  
As a result, we have variable-lengths of object sets residing in a high-dimensional space that spans across time. 
Our objective is to efficiently detect higher-order interactions from these rich yet unordered object representation sets across time. 

In the simplest setting, an interaction between objects in the scene can be represented via summation operation of individual object information.
For example, one method is to add the learnable representations and project these representations into a high-dimensional space where the object interactions can be exploited by simply summing up the object representations. 
Another approach which has been widely used with images is by pairing all possible object candidates (or subject-object pairs)~\cite{chao2017learning, dai2017detecting,hu2016modeling,santoro2017simple,zhang2017visual}. 
However, this is infeasible for video, since a video typically contains hundreds or thousands of frame and the set of object-object pairs is too large to fully represent. 
Detecting object relationships frame by frame is computationally expensive, and the temporal reasoning of object interactions is not used.

\begin{figure}[t]
    \centering
    \includegraphics[width=0.48\textwidth]{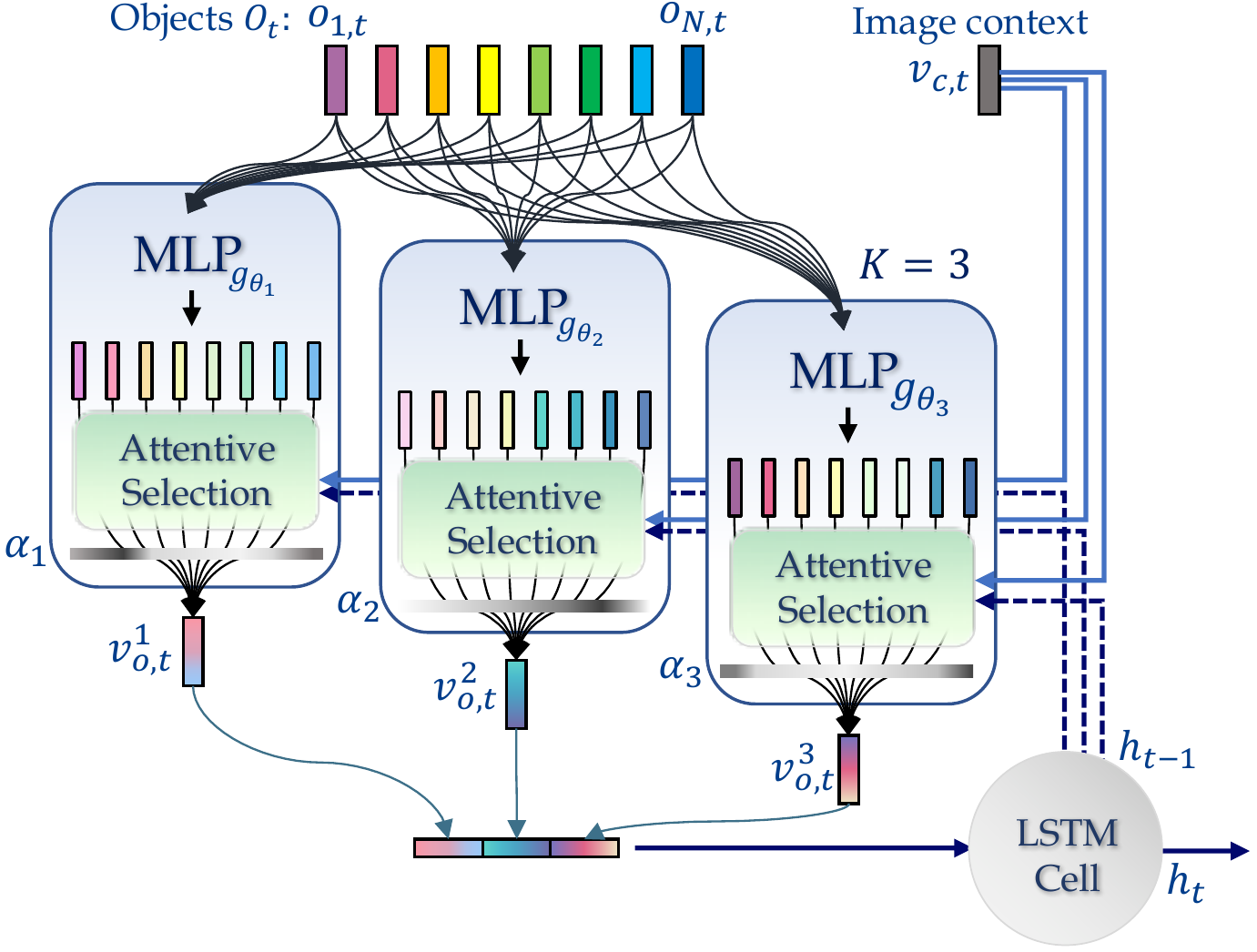}
    \caption{
    \textbf{Recurrent Higher-Order Interaction} module
    dynamically selects $K$ groups of arbitrary objects with detected inter-object relationships via learnable attention mechanism.
    This attentive selection module uses the overall image context representation $v_{c,t}$, current set of (projected) objects $O_t$, and previous object interactions $h_{t-1}$ to generate $k^{th}$ weights $\alpha_k$ for $k^{th}$ selections.
    The higher-order interaction between groups of selected objects is then modeled via concatenation and the following LSTM cell.
    }
    \label{fig:recurrent-hoi}
    \vspace{-0.2in}
\end{figure}

\textbf{Recurrent Higher-Order Interaction (HOI):}
\label{sec:Attentive selection}
To overcome these issues, we propose a generic recurrent module for detecting higher-order object interactions for fine-grained video understanding problems, as shown in Figure~\ref{fig:recurrent-hoi}. 
The proposed recurrent module dynamically selects object candidates which are important to discriminate the human actions. The combinations of these objects are then concatenated to model higher order interaction using group to group or triplet groups of objects.

First, we introduce learnable parameters for the incoming object features via MLP projection $g_{\theta_k}$, since the object features are pre-trained from another domain and may not necessarily present interactions towards action recognition. 
The projected object features are then combined with overall image content and previous object interaction to generate $K$ sets of weights to select $K$ groups of objects~\footnote{The number $K$ depends on the complexity of the visual scene and the requirement of the task (in this case, action recognition). We leave dynamically selecting $K$ to future work.}. 
Objects with inter-relationships are selected from an attention weight, which generates a probability distribution over all object candidates. 
The attention is computed using inputs from current (projected) object features, overall image visual representation, and previously discovered object interactions (see Figure~\ref{fig:recurrent-hoi}), which provide the attention mechanism with maximum context. 
\begin{equation}\label{eq:attention}
\alpha_{k} = Attention(g_{\theta_k} (O_t), v_{c,t}, h_{t-1})
\end{equation}
where the input $O_t$ is a set of \textit{objects}: $O_t = \big\{o_{1, t}, o_{2, t}, ..., o_{N, t}\big\}, o_{n, t} \in \mathbb{R}^m$ is the $n^{th}$ object feature representation at time $t$.
The $g_{\theta_k}$ is a MLP with parameter $\theta_k$, the parameters are learnable synaptic weights shared across all objects $o_{n,t}$ and through time $t$. 
$v_{c,t}$ denotes as encoded image feature at current time $t$, and $h_{t-1}$ is the previous output of LSTM cell which represents the previous discovered object interaction. 
Formally, given an input sequence, a LSTM network computes the hidden vector sequences $\mathbf{h} = \big( h_1, h_2, ..., h_T\big)$.  
Lastly, $\alpha_{k}$ is an attention weight computed from the proposed attention module. 

% The attended object feature at time $t$ is calculated as a convex combination of all object candidates:
% \begin{equation}
% v_{{o, t}}^k=\sum_{i}^{} \alpha_{k_i} (g_{\theta_k}  (o_{i, t}))
% \end{equation}
% where the output $v_{{o, t}}^k$ is a single feature vector representation which encodes the
% $k^{th}$ object inter-relationships of a video frame at time $t$, and $k$ ranges from $\big\{1, 2, ..., K\big\}$ representing the number of groups for inter-relationships. 
% For example, $K=3$ in Figures~\ref{fig:higher-order-interaction} and \ref{fig:recurrent-hoi}.

\begin{figure}[t]
    \centering
    \includegraphics[width=0.45\textwidth]{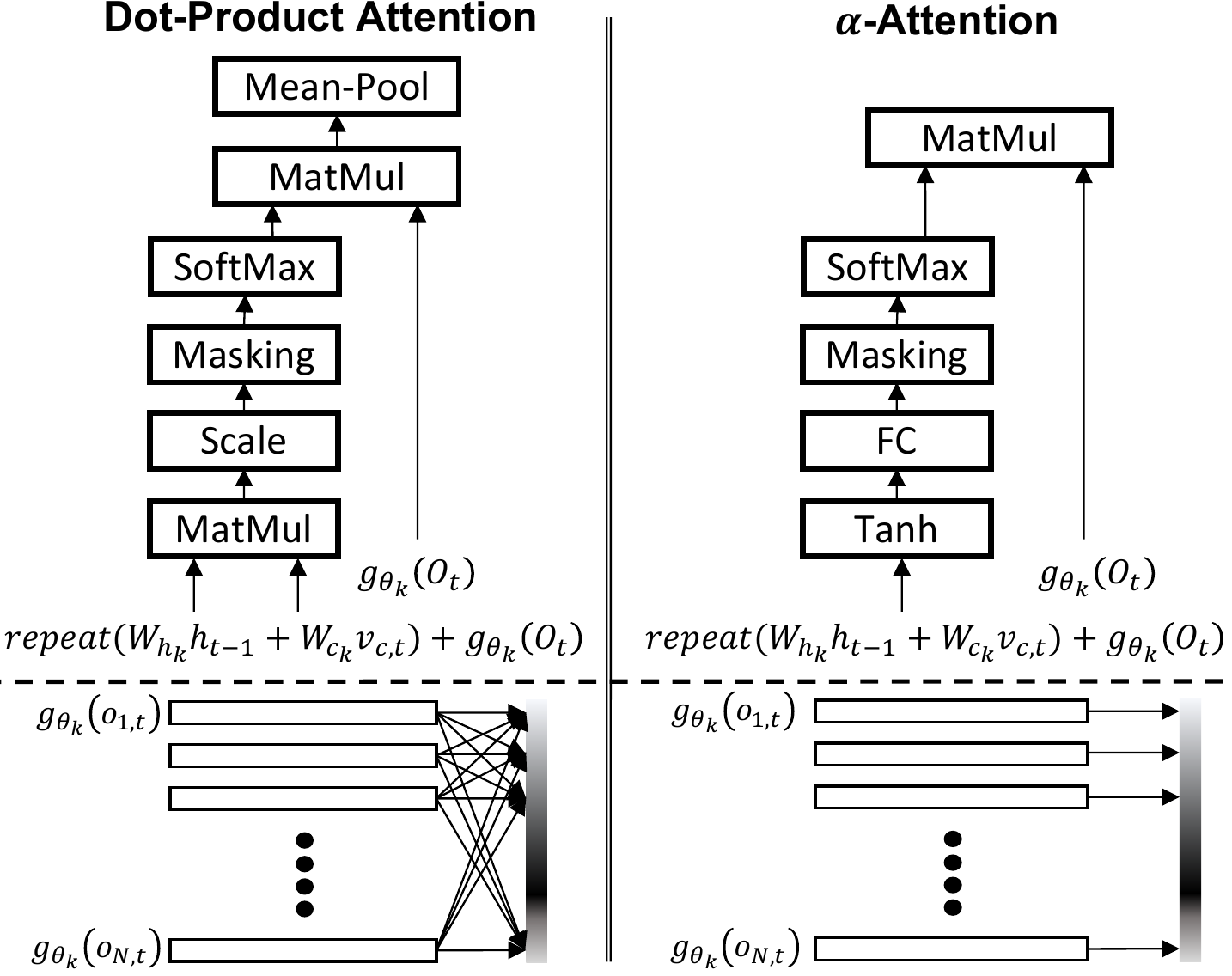}
    \caption{
    Attention modules: dot-product attention and $\alpha$-attention. Both attention mechanisms take input from overall image representation $v_{c,t}$, current set of objects $O_t$, and previous object interactions $h_{t-1}$ computed from LSTM cell at time $t-1$.
    }
    \label{fig:attn}
    \vspace{-0.1in}
\end{figure}

\textbf{Attentive selection module:}
% As shown in Eq. \ref{eq:attention}, our selection module takes three inputs and output a probability distribution over the objects, which will be later used for detecting higher-order interactions. 
Here we discuss two possible choices for the attention module, as shown in Figure~\ref{fig:attn}. Dot-product attention considers inter-relationships when selecting the objects, and $\alpha$-attention does not. 

\textbf{- Dot-product attention:}
In order to model higher-order interactions, which models inter-object relationships in each group of selected objects, we use dot-product attention since the attention weights computed for each object is the combination of all objects. 

Formally, the current image representation $v_{c,t}$ and the last object interaction representation $h_{t-1}$ are first projected to introduce learnable weights. The projected $v_{c,t}$ and $h_{t-1}$ are then repeated and expanded $N$ times (the number of objects in $O_t$). We directly combine this information with projected objects via matrix addition and use it as input to dot-product attention.  
We added a scale factor as in \cite{vaswani2017attention}.
The input to the first matrix multiplication and the attention weights over all objects can be defined as: 
\begin{equation}
{X_{k}} = repeat(W_{h_k} h_{t-1} + W_{c_k} v_{c,t}) + g_{\theta_k} (O_t)
\end{equation}
\begin{equation}
% \alpha_{k} = softmax(\frac{x_{k} \otimes x_{k}}{\sqrt{d_{\theta}}})
\alpha_{k} = softmax(\frac{{X_{k}}^\top {X_{k}} }{\sqrt{d_\theta}})
\end{equation}
% \begin{equation}
% \alpha_{\theta_c} = softmax(\frac{W_q v_{c, t} \times {W_q v_{c, t}}^T}{\sqrt{n}}) W_q v_{c, t}
% \end{equation}
% where $W_{h_k} \in \mathbb{R}^{d_{g_{\theta_k}} \times d_h}$ and $W_{c_k} \in \mathbb{R}^{d_{g_{\theta_k}} \times d_{v_{c,t}}}$ are learned weights for $h_{t-1}$ and $v_{c,t}$. 
where $W_{h_k} \in \mathbb{R}^{d_{\theta} \times d_h}$ and $W_{c_k} \in \mathbb{R}^{d_{\theta} \times d_{v_{c,t}}}$ are learned weights for $h_{t-1}$ and $v_{c,t}$,
$d_{\theta}$ is the dimension of last fully-connected layer of $g_{\theta_k}$,
% $x_{k}$ is the input to $k^{th}$ attention module,
${X_{k}} \in \mathbb{R}^{d_{\theta} \times N }$ is the input to $k^{th}$ attention module,
and $\sqrt{d_{\theta}}$ is a scaling factor, 
$\alpha_{k} \in \mathbb{R}^{N \times N }$ is the computed $k^{th}$ attention.
We omit the bias term for simplicity. 
The attended object feature at time $t$ is then calculated as mean-pooling on weighted objects:
\begin{equation}
v_{{o, t}}^k = \overline{\alpha_k\ (g_{\theta_k} (O_t))^\top}
\end{equation}
where the output $v_{{o, t}}^k$ is a single feature vector representation which encodes the $k^{th}$ object inter-relationships of a video frame at time $t$.

\textbf{- $\alpha$-attention:}
The $\alpha$-attention uses the same input format as dot-product attention, but the attention is computed using a \textit{tanh} function and a fully-connected layer: 
\begin{equation}
\alpha_{k} = softmax({w_k}^\top tanh({X_{k}}))
\end{equation}
% where $w_k \in \mathbb{R}^{d_{x_k}}$ is learned weights. 
where $w_k \in \mathbb{R}^{d_{\theta}}$ is a learned weight, and $\alpha_{k} \in \mathbb{R}^{1 \times N }$ is the computed $k^{th}$ attention.
% Note that the dimension of $\alpha_{k}$ is different from dot-product attention. 
The attended object feature at time $t$ is then calculated as a convex combination:
\begin{equation}
v_{{o, t}}^k=\sum_{n}^{} \alpha_{k_n} (g_{\theta_k}  (o_{n, t}))
\end{equation}

We use the $\alpha$-attention as a baseline to show how considering the inter-relationships of objects (dot-product attention) can further improve the accuracy when ROIs are selected separately.

Finally, for both attention mechanisms, the selected object candidates $v_{{o, t}}^k$ are then concatenated and used as the input to a LSTM cell.
The output $v_{oi,t}$ is then defined as the higher-order object interaction representation at current time $t$.
\begin{equation}
v_{oi,t} = LSTMCell (v_{{o, t}}^1 \Vert v_{{o, t}}^2 \Vert ... \Vert v_{{o, t}}^K)
\end{equation}
where $\Vert$ denotes concatenation between feature vectors. The last hidden state of the LSTM cell $h_{T}=v_{oi,T}$ is the representation of overall object interactions for the entire video sequence.

Note that by concatenating selected inter-object relationships into a single higher-order interaction representation, the selective attention module tends to select different groups of inter-relationships, since concatenating duplicate inter-relationships does not provide extra information and will be penalized. 
% Therefore, our method forces the attention module to focus on different inter-object relationships. 
% For an analysis of what inter-relationships are selected, please refer to the supplement. 
For an analysis of what inter-relationships are selected, please refer to Sec.~\ref{sec:qualitative-kinetics}.

\subsubsection{Late fusion of coarse and fine}
Finally, the attended context information $v_{c}$ obtained from the image representation provides coarse-grained understanding of the video, and the object interactions discovered through the video sequences $v_{oi, T}$ provide fine-grained understanding of the video.
We concatenate them as the input to the last fully-connected layer, and train the model jointly to make a final action prediction. 
 
\begin{equation}
p(y) = softmax(W_p (v_{c} \Vert v_{oi,T}) + b_p)
\end{equation}
where $W_p \in \mathbb{R}^{d_{y} \times (d_{v_c} + d_{v_{oi,T}})}$ and $b_p \in \mathbb{R}^{d_y}$ are learned weights and biases.

\subsection{Video Captioning Model}
We now describe how \netshort\ can be extended from sequence-to-one to a sequence-to-sequence problem for video captioning. 
Our goal in providing fine-grained information for video captioning is that, for each prediction of the word, the model is aware of the past generated word, previous output, and the summary of the video content. 
At each word generation, it has the ability to selectively attend to various parts of the video content in both space and time, as well as to the detected object interactions. 

\begin{figure}[t]
    \centering
    \includegraphics[width=0.48\textwidth]{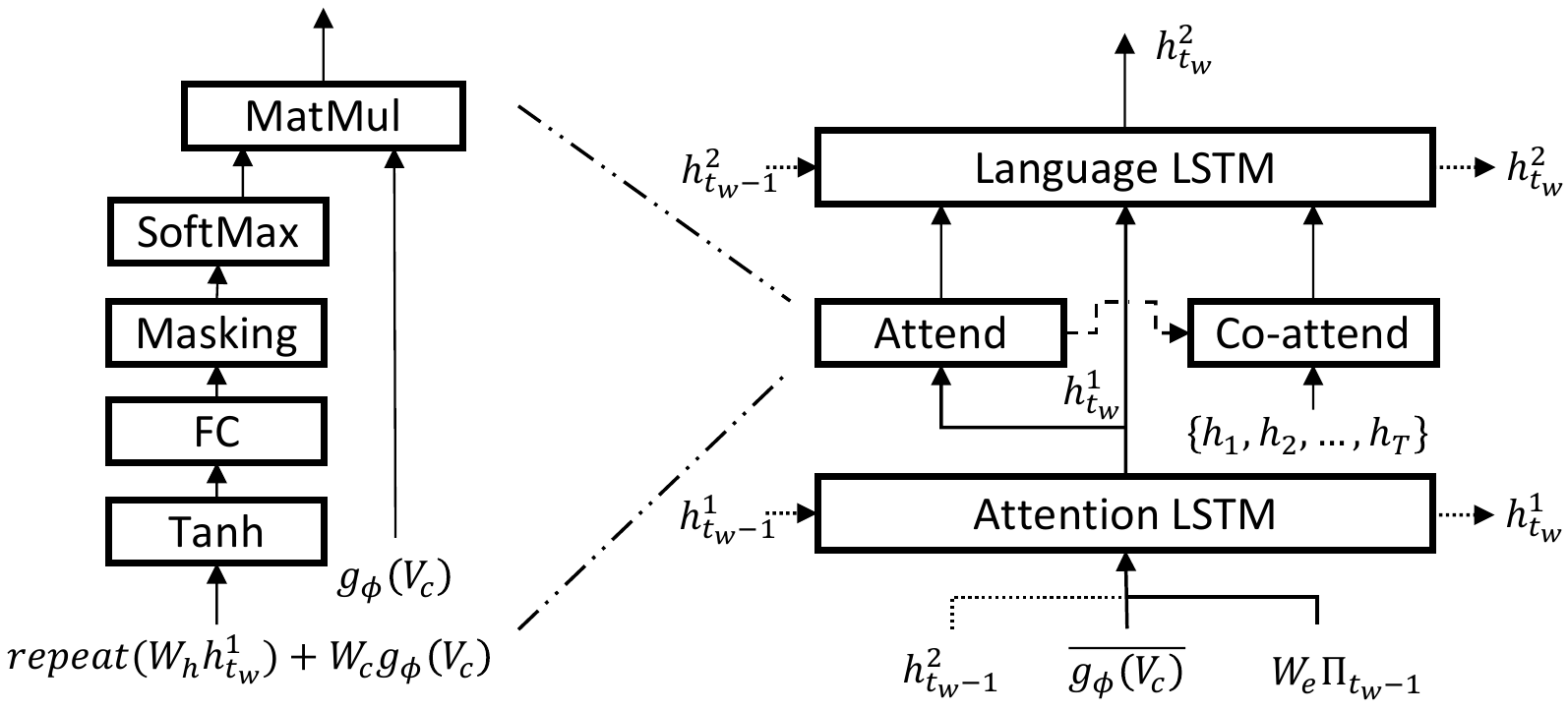}
    \caption{
    Overview of the proposed \netshort-Caption for video captioning. 
    The Attention LSTM with $\alpha$-attention is used to selectively attend to temporal video frame features. 
    The computed temporal attention is then used to attend to temporal object interactions $\{ h_1, h_2, ..., h_{T} \}$ (see Figure~\ref{fig:recurrent-hoi}). 
    Concatenation of the outputs of Attention LSTM, attended video frame feature, and attended object interactions is then used as input for language decoder LSTM.
    }
    \label{fig:model-caption}
    \vspace{-0.1in}
\end{figure}

Our \textbf{\netshort-Caption} is inspired by prior work using hierarchical LSTM for captioning tasks~\cite{anderson2017bottom,song2017hierarchical}, and we extend and integrate it with \netshort\ so that the model can leverage the detected higher-order object interactions. 
We use a two-layered LSTM integrated with the coarse- and fine-grained information, as shown in Figure~\ref{fig:model-caption}. 
The two LSTM layers are: Attention LSTM and Language LSTM.
The Attention LSTM identifies which part of the video in spatiotemporal feature space is needed for Language LSTM to generate the next word. 
Different from prior work, which applied attention directly over all image patches in the entire video~\cite{yu2016video}, i.e. attended to objects individually, our attentive selection module attends to object interactions while considering their temporal order. 

\textbf{Attention LSTM:}
The Attention LSTM fuses the previous hidden state output of Language LSTM $h_{t_w-1}^2$, overall representation of the video, and the input word at time $t_w-1$ to generate the hidden representation for the following attention module.
Formally, the input to Attention LSTM can be defined as:

\begin{equation}
x_{t_w}^1 = h_{t_w-1}^2 \ \Vert \ \overline{g_{\phi} (V_{c}}) \ \Vert \ W_e \Pi_{t_w-1}
\end{equation}
% where $h_{t-1}^2$ is the previous output of Language LSTM,
where $\overline{g_{\phi}(V_{c})}$ is the projected and mean-pooled image features,
$g_\phi$ is a MLP with parameters $\phi$,
$W_e \in \mathbb{R}^{E \times \Sigma}$ is a word embedding matrix for a vocabulary of size $\Sigma$,
and $\Pi_{t_w-1}$ is one-hot encoding of the input word at time $t_w-1$.
Note that $t$ is the video time, and $t_w$ is the timestep for each word generation.

\textbf{Temporal attention module:}
We adapt the same $\alpha$-attention module as shown in Figure~\ref{fig:attn} to attend over projected image features $g_{\phi} (V_{c})$. 
The two types of input for this temporal attention module are from outputs of the Attention LSTM and projected image features. 
\begin{equation}
{X_{a}} = repeat(W_h h_{t_w}^1) + W_c g_{\phi} (V_{c})
\end{equation}
where $h_{t_w}^1$ is the output of Attention LSTM,
$W_h \in \mathbb{R}^{d_{\phi} \times d_{h_{t_w}^1}}$ and $W_c \in \mathbb{R}^{d_{\phi} \times d_{\phi}}$ are learned weights for $h_{t_w}^1$ and $g_{\phi} (V_{c})$. 
$d_{\phi}$ is the dimension of the last FC layer of $g_{\phi}$.

\textbf{Co-attention:}
We directly apply the temporal attention obtained from image features on object interaction representations $\mathbf{h} = \big( h_1, h_2, ..., h_T\big)$ (see Sec~\ref{sec:obj-interaction} for details).

\textbf{Language LSTM:}
Finally, the Language LSTM takes in input which is the concatenation of output of the Attention LSTM $h_{t_w}^1$, attended video representation $\Myhat{v_{c, t_w}}$, and co-attended object interactions $\Myhat{h_{t_w}}$ at timestep $t_w$. 

\begin{equation}
x_{t_w}^2 = h_{t_w}^1 \ \| \ \Myhat{v_{c, t_w}} \ \Vert \ \Myhat{h_{t_w}}
\end{equation}

The output of Language LSTM is then used to generate each word, which is a conditional probability distribution defined as:
\begin{equation}
p(y_{t_w} | y_{1:t_w-1}) = softmax(W_p h_{t_w}^2)
\end{equation}
where $y_{1:t_w-1}$ is a sequence of outputs $(y_1, ..., y_{t_w-1})$ and $W_p \in \mathbb{R}^{\Sigma \times d_{h_{t_w}^2}}$ is learned weights for $h_{t_w}^2$. 
All bias terms are omitted for simplicity.

\section{Datasets and Implementations} 
\subsection{Datasets:}
% Video understanding can be generally classified into two tasks: sequence-to-one and sequence-to-sequence.
% In the following, we first describe the datasets used for these two problems followed by the implementation details and training procedures of our \netshort. 
% Code is developed using the PyTorch deep learning framework. 

\textbf{Kinetics dataset:}
To evaluate \netshort\ on a sequence-to-one problem for video, we use the Kinetics dataset for action recognition~\cite{kay2017kinetics}. 
The Kinetics dataset contains 400 human action classes and has approximately 300k video clips (833 video hours).
Most importantly, different from previous datasets which mostly cover sports actions~\cite{karpathy2014large, kuehne2011hmdb, soomro2012ucf101}, Kinetics includes human-object interactions and human-human interactions. 
We sampled videos at 1 FPS only, as opposed to sampling at 25 FPS reported for Kinetics~\cite{kay2017kinetics}.

\textbf{ActivityNet Captions dataset:}
To evaluate \netshort-Caption\ on a sequence-to-sequence problem for video, we use ActivityNet Captions for video captioning. The ActivityNet Captions dataset contains 20k videos and has total of 849 video hours with 100K total descriptions. 
% Although the number of videos is less compared to Kinetics, the video hours is in fact slightly longer than Kinetics (approximately 833 hours).
To demonstrate our proposed idea, we focus on providing fine-grained understanding of the video to describe video events with natural language, as opposed to identifying the temporal proposals. We thus use the ground truth temporal segments and treat each temporal segment independently. 
We use this dataset over others because ActivityNet Captions is action-centric, as opposed to object-centric~\cite{krishna2017dense}. 
This fits our goal of detecting higher-order object interactions for understanding human actions.
% Following the same procedure as in \cite{krishna2017dense}, 
All sentences are capped to be a maximum length of 30 words. 
We sample predictions using beam search of size 5 for captioning. 
While the previous work sample C3D features every 8 frames~\cite{krishna2017dense}, we only sampled video at maximum 1 FPS. Video segments longer than 30 secs. are evenly sampled at maximum 30 samples.

\subsection{Implementation Details:}
We now discuss how to extract image and object features for both Kinetics and ActivityNet Captions. 

\textbf{Image feature:}
We fine-tune a pre-trained ResNeXt-101~\cite{xie2017aggregated} on Kinetics sampled at 1 FPS (approximately 2.5 million images). 
We use SGD with Nesterov momentum as the optimizer. 
The initial learning rate is $1e-4$ and drops by 10x when validation loss saturates for 5 epochs. 
The weight decay is $1e-4$ and the momentum is 0.9, and the batch size is 128. 
We use standard data augmentation by randomly cropping and horizontally flipping video frames during training. 
When extracting image features, the smaller edge of the image is scaled to 256 pixels and we crop the center of the image as input to the fine-tuned ResNeXt-101. 
Each image feature is a 2048-d feature vector. 

\textbf{Object feature:}
We generate the object features by first obtaining the coordinates of ROIs from a Deformable R-FCN~\cite{dai2017deformable} (pre-trained on MS-COCO) with ResNet-101~\cite{he2016deep} as backbone architecture. 
% The Deformable R-FCN was trained on MS-COCO train and validation dataset~\cite{lin2014microsoft}. 
We set the IoU threshold for NMS to be 0.2. 
Empirically, we found that it is important to maintain a balance of image and object features, especially when image features were obtained from a network which was fine-tuned on the target dataset. 
Thus, for each of the ROIs, we extract features using coordinates and adaptive max-pooling from the same model (ResNeXt-101) that was fine-tuned on Kinetics. 
% Using the ROI features obtained from Deformable R-FCN only has minor affect when trained with image features. 
The resulting object feature for each ROI is a 2048-d feature vector. 
ROIs are ranked according to their ROI scores. We select top 30 objects for Kinetics and top 15 for ActivityNet Captions. 
Note that we have a varied number of ROIs for each video frame, and video length can also be different.  
We do not use the object class information since we may miss some of the objects that were not detected, due to the cross-domain problem.
For the same reason, the bounding-box regression process is not performed here since we do not have the ground-truth bounding boxes.

\textbf{Training:}
We train \netshort\ and \netshort-Caption\ with ADAM optimizer. 
The initial learning rate is set to $1e-5$ for Kinetics and $1e-3$ for ActivityNet Captions.
Both learning rates automatically drop by 10x when validation loss is saturated. 
The batch sizes are $64$ and $32$ respectively for Kinetics and ActivityNet Captions.

\begin{table}[t]
    \centering
    \caption{
        Prediction accuracy on the Kinetics validation set.
        % \textit{Obj feature} use only the extracted object features.
        All of our results use only RGB videos sampled at 1 FPS.
        Maximum number of objects per frame is set to be 30.
        }
    \label{table:kinetics-eval}
    \small
    \begin{tabular}{ccc}
    Method                               & Top-1 			& Top-5 		\\ \hline
    I3D\footnotemark (25 FPS)~\cite{carreira2017quo} (test)       & 71.1  			& 89.3  		\\
    % I3D~\cite{carreira2017quo} (test)    Flow    & 63.4  			& 84.9  		\\
    % I3D~\cite{carreira2017quo} (test)    RGB + Flow & 74.2  			& 91.3  		\\ \hline
    TSN (Inception-ResNet-v2) (2.5 FPS)~\cite{bian2017revisiting, WangXWQLTV16}    & 73.0  & 90.9          \\ \hline
    % Shifting Attention Network (Inception-ResNet-v2)~\cite{bian2017revisiting}  RGB + Flow + Audio  & 77.7  & 93.2          \\ \hline
    Ours (1 FPS) \\ \hline
    Img feature + LSTM (baseline)                          & 70.6  			& 89.1  		\\
    Img feature + temporal SDP-Attention                 & 71.1  			& 89.6  		\\
    Obj feature (mean-pooling)                      & 72.2  		    & 90.2  		\\
    Img + obj feature (mean-pooling)                     & 73.1  		    & 91.1  		\\
    % \netshort\ ($\alpha$-attention) w/o img context            & 73.8              & 91.4          \\
    \netshort\ ($\alpha$-attention)            & 73.9              & 91.5          \\
    % \netshort\ (dot-product attention) w/o img context            & 74.0              & 91.4          \\
    \netshort\ (dot-product attention)              & \textbf{74.2}     & \textbf{91.7}
    % \netshort\ (three groups)              & \textbf{74.1}     & \textbf{91.5} 
    \end{tabular}
    \vspace{-0.1in}
\end{table}
% \footnotetext{Results obtained from DeepMind's \href{https://github.com/deepmind/kinetics-i3d}{GitHub repository}}
\footnotetext{Results obtained from \href{https://github.com/deepmind/kinetics-i3d}{https://github.com/deepmind/kinetics-i3d}}

\section{Evaluation}
\label{sec:evaluation}
\subsection{Action recognition on Kinetics:}
\label{sec:evaluation-action}
In this section, we conduct an ablation study of \netshort\ on Kinetics.

\textbf{Does temporal SDP-Attention help?}
Several studies have pointed out that using temporal mean-pooling or LSTMs may not be the best method to aggregate the sequence of image representations for videos~\cite{bian2017revisiting,ma2017ts,miech2017learnable}. To overcome this issue, we use temporal SDP-Attention instead of LSTM. 
As we can see from Table~\ref{table:kinetics-eval}, using temporal SDP-Attention has proven to be superior to traditional LSTM and already performs comparably with 3D ConvNet that uses a much higher video sampling rate. 

\textbf{Does object interaction help?}
We first evaluate how much higher-order object interactions can help in identifying human actions. 
Considering mean-pooling over the object features to be the simplest form of object interaction, we show that mean-pooling over the object features per frame and using LSTM for temporal reasoning has already outperformed single compact image representations, which is currently the trend for video classification methods. 
% Although different methods have proposed to aggregate temporal representations, we typically use single compact representation for each frame.
Directly combining image features with temporal SDP-Attention and object features over LSTM further reaches 73.1\% top-1 accuracy. 
This already outperforms the state-of-the-art TSN~\cite{WangXWQLTV16} method using a deeper ConvNet with a higher video sampling rate. 
Beyond using mean-pooling as the simplest form of object interaction, our proposed method to dynamically discover and model higher-order object interactions further achieved 74.2\% top-1 and 91.7\% top-5 accuracy.
The selection module with dot-product attention, in which we exploit the inter-relationships between objects within the same group, outperforms $\alpha$-attention where the inter-relationships are ignored. 

\begin{table}[t]
    \centering
    \caption{
        Comparison of pairwise (or triplet) object interaction with the proposed higher-order object interaction with dot-product attentive selection method on Kinetics. 
        The maximum number of objects is set to be 15. 
        % FLOP is calculated per video. For details on calculating FLOP, please refer to the supplementary material. 
        FLOP is calculated per video. For details on calculating FLOP, please refer to Sec.~\ref{sec:sub-FLOPs}. 
        }
    \label{table:kinetics-selection}
    \small
    \begin{tabular}{cccc}
    Method                          & Top-1 			& Top-5 	& FLOP ($e^9$) \\ \hline
    % \multicolumn{4}{c}{max number of objs = $15$} \\ \hline
    Obj (mean-pooling)                             & 73.1  		    & 90.8 		& ~1.9  \\ 
    Obj pairs (mean-pooling)                       & 73.4 		        & 90.8  	& ~18.3  \\
    Obj triplet (mean-pooling)                     & 72.9 		        & 90.7 		& ~77.0  \\ \hline
    \netshort\ ($K=1$)              & 73.9              & 91.3         & ~2.7 \\
    \netshort\ ($K=2$)              & 74.2              & 91.5         & ~5.3 \\
    \netshort\ ($K=3$)              & \textbf{74.2}              & \textbf{91.7}         & ~8.0 \\
    % \multicolumn{4}{c}{max number of objs = $30$} \\ \hline
    % \netshort\ ($K=1$)              & 74.0              & 91.4         & ? \\
    % \netshort\ ($K=2$)              & 74.2              & 91.5         & ? \\
    % \netshort\ ($K=3$)              & \textbf{74.2}        & \textbf{91.7}         & ?
    \end{tabular}
    \vspace{-0.1in}
\end{table}

\begin{table*}[t]
    \centering
    \caption{
        METEOR, ROUGE-L, CIDEr-D, and BLEU@N scores on the ActivityNet Captions test and validation set. All methods use ground truth proposal except LSTM-A$_3$~\cite{ghanem2017activitynet}.
        Our results with ResNeXt spatial features use videos sampled at maximum 1 FPS only.
        }
    \label{table:caption-eval}
    \small
    \scalebox{0.85}{
    \begin{tabular}{lccccccc}
    Method  & B@1   & B@2  & B@3 & B@4    & ROUGE-L    & METEOR    & CIDEr-D   \\ \hline 
    \multicolumn{1}{c}{\textbf{Test set}} & & & & & & & \\ \hline
    LSTM-YT~\cite{venugopalan2014translating} (C3D)  & 18.22    & 7.43  & 3.24  & 1.24      & -     & 6.56   &14.86 \\
    S2VT~\cite{venugopalan2015sequence} (C3D)  & 20.35    & 8.99  & 4.60  & 2.62      & -     & 7.85   & 20.97 \\
    H-RNN~\cite{yu2016video} (C3D)  & 19.46    & 8.78  & 4.34  & 2.53      & -     & 8.02   & 20.18 \\
    S2VT + full context~\cite{krishna2017dense} (C3D)  & 26.45    & 13.48  & 7.21  & 3.98      & -     & 9.46   & 24.56 \\
    \multirow{2}{*}{\begin{tabular}[c]{@{}c@{}}LSTM-A$_3$ + policy gradient + retrieval~\cite{ghanem2017activitynet}\\ (ResNet + P3D ResNet~\cite{Qiu_2017_ICCV})\end{tabular}} & \multirow{2}{*}{-} & \multirow{2}{*}{-} & \multirow{2}{*}{-} & \multirow{2}{*}{-} & \multirow{2}{*}{-} & \multirow{2}{*}{12.84} & \multirow{2}{*}{-} \\
    &   &   &   &   &   &   &   \\ \hline 
    \multicolumn{1}{c}{\textbf{Validation set (Avg. 1st and 2nd)}} & & & & & & & \\ \hline
    %     % LSTM-YT~\cite{venugopalan2014translating} (C3D)         & 16.21     & 7.22     & 3.04    & 1.32    & 18.43    & 7.81    & 31.71	\\
    %     % LSTM-YT~\cite{venugopalan2014translating} (Kinetics)    & 17.37     & 8.23     & 3.68    & 1.65    & 19.14    & 8.62 	& 36.03	\\ 
    LSTM-A$_3$ (ResNet + P3D ResNet)~\cite{ghanem2017activitynet}   & 17.5 & 9.62 & 5.54    & \textbf{3.38}  & 13.27 & 7.71  & 16.08 \\
    \multirow{2}{*}{\begin{tabular}[c]{@{}c@{}}LSTM-A$_3$ + policy gradient + retrieval~\cite{ghanem2017activitynet}\\ (ResNet + P3D ResNet~\cite{Qiu_2017_ICCV})\end{tabular}} & \multirow{2}{*}{17.27} & \multirow{2}{*}{9.70} & \multirow{2}{*}{5.39} & \multirow{2}{*}{3.13} & \multirow{2}{*}{14.29} & \multirow{2}{*}{8.73} & \multirow{2}{*}{14.75} \\
    &   &   &   &   &   &   &   \\ 
    % ================= tIoU = 0.3
    % \netshort-Caption w/o obj (C3D)                         & 17.21 & 8.02  & 3.55  & 1.47  & 18.81 & 8.45  & 36.90 \\
    % \netshort-Caption w/o obj (ResNeXt)                     & 18.58 & 9.19  & 4.20  & 1.80  & 20.31 & 9.44  & 41.40 \\
    % \netshort-Caption + obj w/o co-attention (ResNeXt)      & \textbf{19.85} & \textbf{9.79}  & \textbf{4.52}  & 2.02  & \textbf{21.00} & \textbf{9.78}  & \textbf{43.28} \\
    % \netshort-Caption + obj w/ co-attention (ResNeXt)     & \textbf{19.59}    & \textbf{9.77}   & \textbf{4.49}   & 2.01 & \textbf{20.99}    & \textbf{9.76} & \textbf{43.66} 
    % ================= tIoU = 0.9
    \netshort-Caption | img (C3D)                         & 17.18 & 7.99  & 3.53  & 1.47  & 18.78 & 8.44  & 38.22 \\
    \netshort-Caption | img (ResNeXt)                     & 18.81 & 9.31  & 4.27  & 1.84  & 20.46 & 9.56  & 43.12 \\
    \netshort-Caption | obj (ResNeXt)                     & 19.07 & 9.48  & 4.38  & 1.92  & 20.67 & 9.56  & 44.02 \\
    \netshort-Caption | img + obj | no co-attention (ResNeXt)      & \textbf{19.93} & 9.82  & \textbf{4.52}  & 2.03  & 21.08 & 9.79  & 44.81 \\
    \netshort-Caption | img + obj | co-attention (ResNeXt)     & 19.78    & \textbf{9.89}   & \textbf{4.52}   & 1.98 & \textbf{21.25}    & \textbf{9.84} & \textbf{44.84}
\end{tabular}
}
\vspace{-0.1in}
\end{table*}

\textbf{Does attentive selection help?}
Prior work on visual relationships and VQA concatenate pairwise object features for detecting object relationships.
In this experiment, we compare the traditional way of creating object pairs or triplets with our proposed attentive selection method.
We use temporal SDP-Attention for image features, and dot-project attention for selecting object interactions.
% , and the maximum number of objects is set to be 15 per frame for this experiment. 
As shown in Table~\ref{table:kinetics-selection}, concatenating pairwise features marginally improves over the simplest form of object interactions while increasing the computational cost drastically.
% with the expense of computational resources.
By further concatenating three object features, the space for meaningful object interactions becomes so sparse that it instead reduced the prediction accuracy, and the number of operations (FLOP) further increases drastically.
On the other hand, our attentive selection method can improve upon these methods while saving significant computation time.
Empirically, we also found that reducing the number of objects per frame from 30 to 15 yields no substantial difference on prediction accuracy.
This indicates that the top 15 objects with highest ROI score are sufficient to represent fine-grained details of the video.
% For detailed qualitative analysis of how objects are selected at each timestep and how \netshort\ reasons over a sequence of object interactions, please see the supplement.
For detailed qualitative analysis of how objects are selected at each timestep and how \netshort\ reasons over a sequence of object interactions, please see Sec.~\ref{sec:qualitative-kinetics}.

We are aware of that integrating optical flow or audio information with RGB video can further improve the action recognition accuracy~\cite{bian2017revisiting, carreira2017quo}. 
We instead focus on modeling object interactions for understanding video in a fine-grained manner, and we consider other modalities to be complementary to our higher-order object interactions. 

\subsection{Video captioning on ActivityNet Captions:}
We focus on understanding human actions for video captioning rather than on temporal proposals. 
Hence, we use ground truth temporal proposals for segmenting the videos and treat each video segment independently. 
All methods in Table~\ref{table:caption-eval} use ground truth temporal proposal, except LSTM-A$_3$~\cite{ghanem2017activitynet}. 
Our performances are reported with four language metrics, including BLEU~\cite{papineni2002bleu}, ROUGH-L~\cite{lin2004rouge}, METEOR~\cite{banerjee2005meteor}, and CIDEr-D~\cite{vedantam2015cider}. 
% Scores are averaged from both first and second validation set.

For fair comparison with prior methods using C3D features, we report results with both C3D and ResNeXt spatial features.
Since there is no prior result reported on the validation set, we compare against LSTM-A$_3$~\cite{ghanem2017activitynet} which reports results on the validation and test sets. 
This allows us to indirectly compare with methods reported on the test set. 
As shown in Table~\ref{table:caption-eval}, while LSTM-A$_3$ clearly outperforms other methods on the test set with a large margin, our method shows better results on the validation sets across nearly all language metrics. 
We do not claim our method to be superior to LSTM-A$_3$ because of two fundamental differences. First, they do not rely on ground truth temporal proposals. 
Second, they use features extracted from an ResNet fine-tuned on Kinetics and another P3D ResNet~\cite{Qiu_2017_ICCV} fine-tuned on Sports-1M, whereas we use a ResNeXt-101 fine-tuned on Kinetics sampled at maximum 1 FPS. 
Utilizing more powerful feature representations has been proved to improve the prediction accuracy by a large margin on video tasks. This also corresponds to our experiments with C3D and ResNeXt features, where the proposed method with ResNeXt features perform significantly better than C3D features.

\textbf{Does object interaction help?}
\netshort-Caption without any object interaction has already outperformed prior methods reported on this dataset. 
Additionally, by introducing an efficient selection module for detecting object interactions, \netshort-Caption further improves across nearly all evaluation metrics, with or without co-attention. 
% We also observed that introducing the co-attention from image features constantly shows improvement on the first validation set but having separate temporal attention for object interaction features show better results on second validation set (please see the supplement for results on each validation set).
We also observed that introducing the co-attention from image features constantly shows improvement on the first validation set but having separate temporal attention for object interaction features show better results on second validation set (please see Sec.~\ref{sec:caption-vals-results} for results on each validation set).

\section{Conclusion}
We introduce a computationally efficient fine-grained video understanding approach for discovering higher-order object interactions. 
Our work on large-scale action recognition and video captioning datasets demonstrates that learning higher-order object relationships provides high accuracy over existing methods at low computation costs. 
We achieve state-of-the-art performances on both tasks with only RGB videos sampled at maximum 1 FPS.
% No motion information is used.
\section*{Acknowledgments}
Zsolt Kira was partially supported by the National Science Foundation and National Robotics Initiative (grant \# IIS-1426998).
\section{Supplementary}

\subsection{Qualitative analysis on Kinetics}
\label{sec:qualitative-kinetics}
\begin{figure*}[t]
    \centering
    \includegraphics[width=1\textwidth]{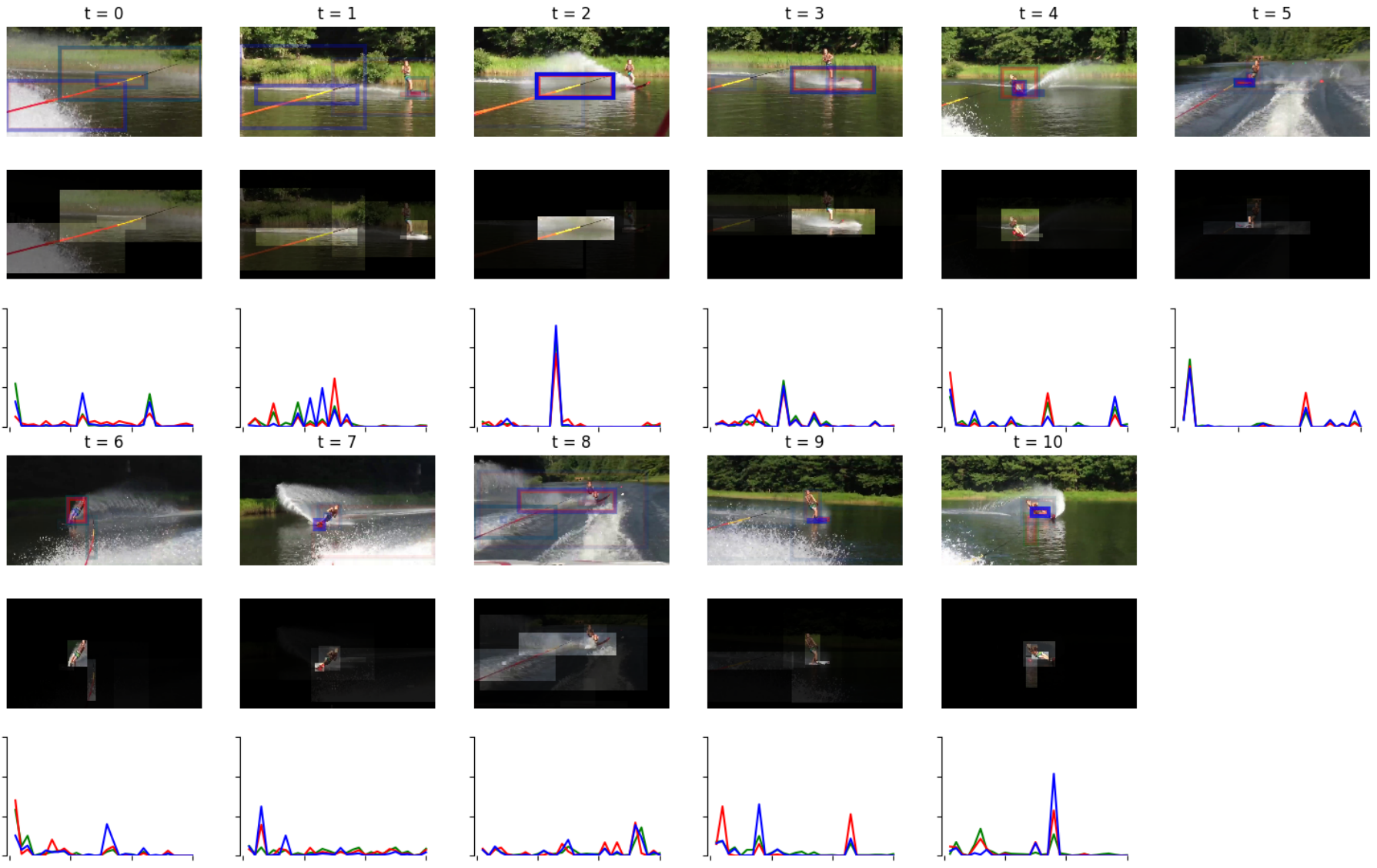}
    \caption{
    \textbf{Water skiing}: Our \netshort\ is able to identify several object relationships and reasons these interactions through time: (1) the rope above the water (2) the wakeboard on the water (3) human riding on the wakeboard (4) rope connecting to the person on the wakeboard. 
    From the distribution of three different attention weights (\textcolor{red}{red}, \textcolor{green}{green}, \textcolor{blue}{blue}), we can also see that the proposed attention method not only is able to select objects with different inter-relationships but also can use a common object to discover different relationships around that object when needed. 
    We observed that our method tends to explore the whole scene at the beginning of the video, and focus on new information that is different from the past. For example, while video frame at first few frames are similar, the model focus on different aspect of the visual representation. 
    }
    \label{fig:water_skiing}
\end{figure*}

\begin{figure*}[t]
    \centering
    \includegraphics[width=1\textwidth]{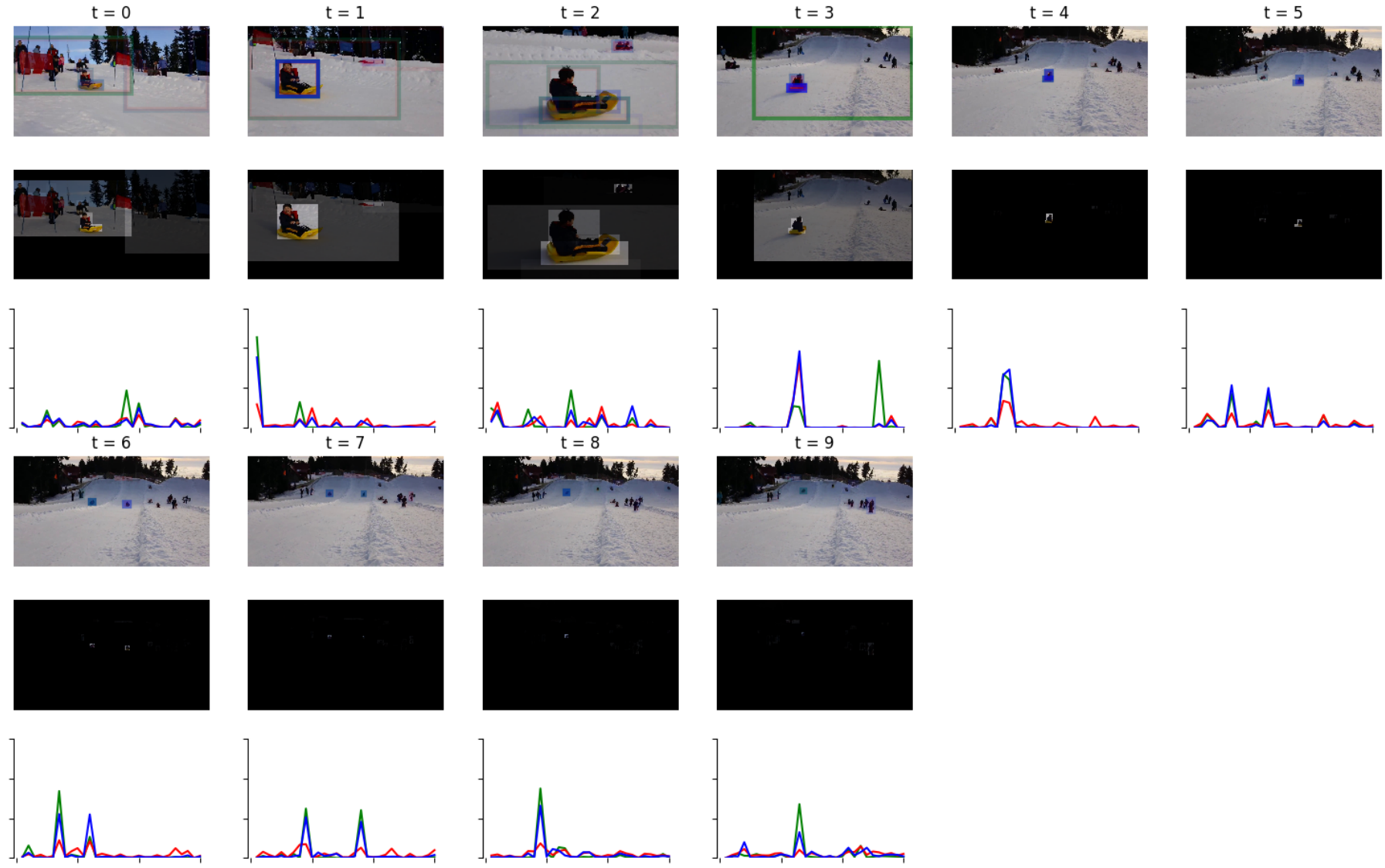}
    \caption{
    \textbf{Tobogganing}: Identifying \textit{Tobogganing} essentially need three elements: toboggan, snow scene, and a human sitting on top. The three key elements are accurately identified and their interaction are highlighted as we can see from $t=1$ to $t=3$. Note that the model is able to continue tracking the person and toboggan throughout the whole video, even though they appear very small towards the end of the video. 
    We can also noticed that our \netshort\ completely ignore the background scene in the last several video frames as they are not informative since they can be easily confused by other 18 action classes involving snow and ice, e.g. \textit{Making snowman}, \textit{Ski jumping}, \textit{Skiing crosscountry}, \textit{Snowboarding}, etc. 
    }
    \label{fig:tobogganing}
\end{figure*}

\begin{figure*}[t]
    \centering
    \includegraphics[width=1\textwidth]{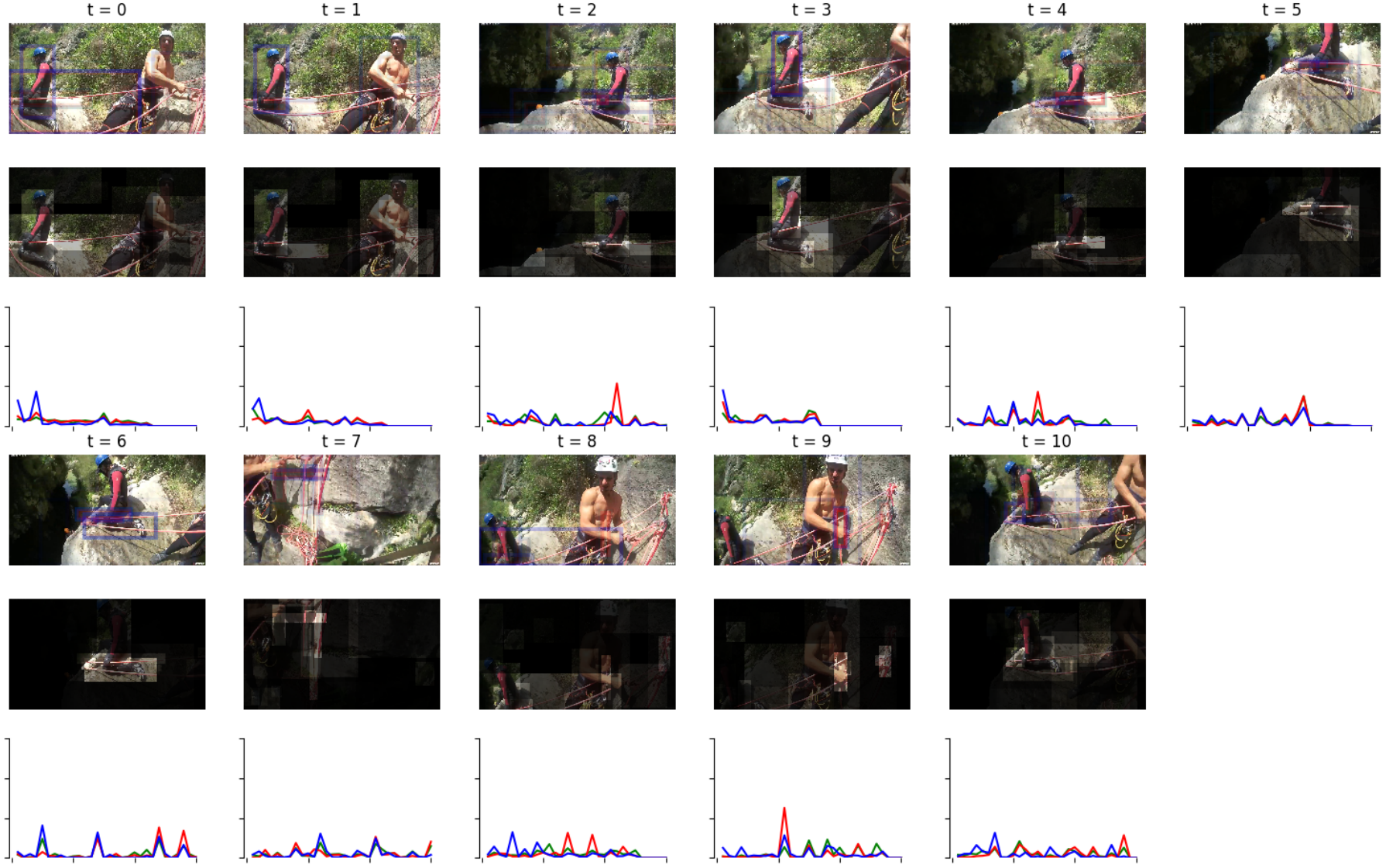}
    \caption{
    \textbf{Abseiling} is challenging since there are similar classes exist: \textit{Climbing a rope}, \textit{Diving cliff}, and \textit{Rock climbing}, which involve ropes, rocks and cliffs. 
    To achieve this, the model progressively identify the interactions and relationships like: human sitting the rock, human holding the rope, and the presence of both rope and rock. 
    This information is proven to be sufficient for predicting \textit{Abseiling} over other ambiguous action classes. 
    }
    \label{fig:abseiling}
\end{figure*}

To further validate the proposed method, we qualitatively show how the \netshort\ selectively attends to various regions with relationships and interactions across time. We show several examples in Figure~\ref{fig:water_skiing}, \ref{fig:tobogganing}, and \ref{fig:abseiling}. 
In each of the figure, the top row of each video frame has generally multiple ROIs with three colors: \textcolor{red}{red}, \textcolor{green}{green}, and \textcolor{blue}{blue}. 
ROIs with the same color indicates that there exist inter-relationships. We then model the interaction between groups of ROIs across different colors.
The color of each bounding box is weighted by the attention generated by the proposed method.  
Thus, if some ROIs are not important, they will have smaller weights and will not be shown on the image.
The same weights are then used to set the transparent ratio for each ROI. The brighter the region is, the more important the ROI is. 

\textbf{Focus on object semantics}
Recent state-of-the-art methods for action recognition rely on single compact representation of the scene. We show that the proposed \netshort\ can focus on the details of the scene and neglect the visual content that maybe irrelevant such as the background information. For example, in Figure~\ref{fig:water_skiing}, the model constantly focus on the rope above the water and the person riding on wakeboard.
% , while neglecting the irrelevant background. 
The same goes for Figure~\ref{fig:tobogganing}. The background scenes with ice and snow are ignored throughout the video since it's ambiguous and easy to be confused with other classes involve snow in the scene.

\textbf{Adjustable inter-relationships selection}
We notice that our \netshort\ tends to explore the whole scene early in the video, i.e. the attentions tend to be distributed to the ROIs that cover large portion of the video frame, and the attentions become more focused after this exploration stage. 

\subsection{Qualitative analysis on ActivityNet Captions}
\label{sec:discussion-Activitynet}
\begin{figure*}[t]
    \centering
    \includegraphics[width=1\textwidth]{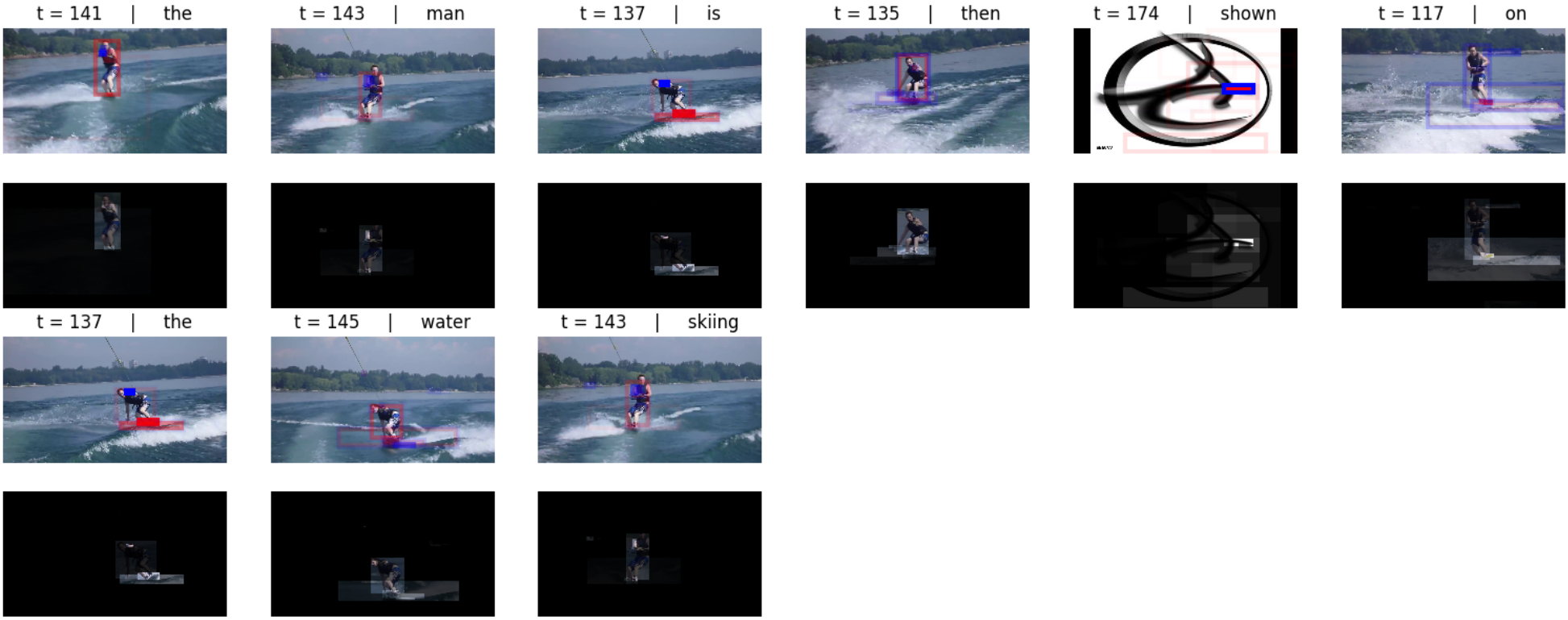}
    \caption{
    \textit{The man is then shown on the water skiing.}
    We can see that the proposed \netshort-Caption often focus on the person and the wakeboard, and most importantly it highlight the interaction between the two, i.e. the person steps on the wakeboard. 
    }
    \label{fig:water_surfing}
    \vspace{-0.1in}
\end{figure*}

\begin{figure*}[t]
    \centering
    \includegraphics[width=1\textwidth]{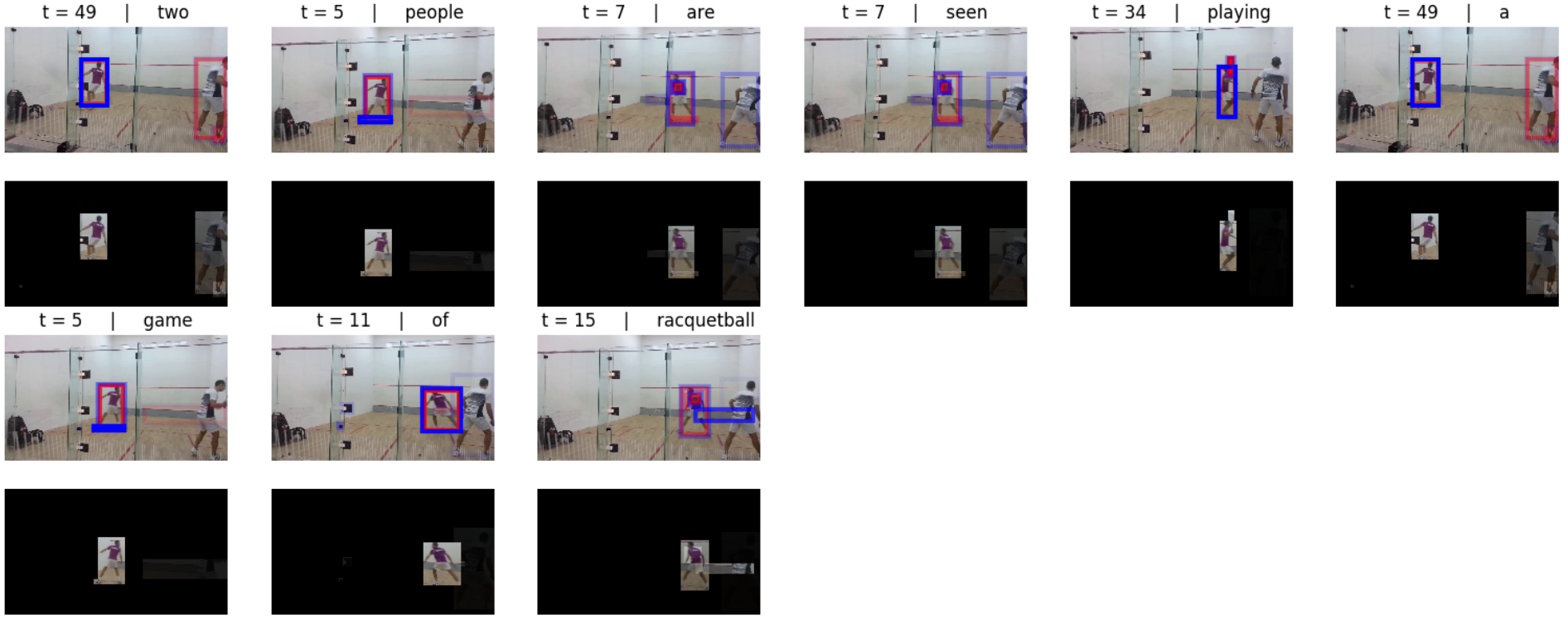}
    \caption{
    \textit{Two people are seen playing a game of racquetball.}
    The \netshort-Caption is able to identify that two persons are playing the racquetball and highlight the corresponding ROIs in the scene. 
    }
    \label{fig:racquetball}
    \vspace{-0.1in}
\end{figure*}

In addition to the qualitative analysis on action recognition task, we now present the analysis on video captioning. 
Several examples are shown in Figure~\ref{fig:water_surfing}, \ref{fig:camel}, and \ref{fig:racquetball}.
At each word generation step, the \netshort-Caption uses the weighted sum of the video frame representations and the weighted sum of object interactions at corresponding timesteps (co-attention). 
Note that, since we aggregate the detected object interactions via the LSTM cell through time, the feature representation of the object interactions at each timestep can be seen as a fusion of interactions at the present and past time. 
Thus, if temporal attention has highest weight on $t=3$, it may actually attend to the interaction aggregated from $t=1$ to $t=3$. 
Nonetheless, we only show the video frame with highest temporal attention for convenience. 
We use \textcolor{red}{red} and \textcolor{blue}{blue} to represent the two selected sets of objects ($K=2$). 

In each of the figures, the video frames (with maximum temporal attention) at different timesteps are shown along with each word generation.
All ROIs in the top or bottom images are weighted with their attention weights. In the top image, ROIs with weighted bounding box edges are shown, whereas, in the bottom image, we set the transparent ratio equal to the weight of each ROI. 
The brighter the region is, the more important the ROI is. 
Therefore, less important ROIs (with smaller attention weights) will disappear in the top image and be completely black in the bottom image. 
When generating a word, we traverse the selection of beam search at each timestep. 

As shown in Figure~\ref{fig:water_surfing}, we can see that the \netshort-Caption can successfully identify the person and the wakeboard. These selections of the two most important objects imply that the person is riding on the wakeboard --- water skiing. 
We also observe that, in Figure~\ref{fig:camel}, the proposed method focuses on the bounding boxes containing both person and the camel. Suggesting that this is a video for people sitting on a camel. However, it failed to identify that there are in fact multiple people in the scene and there are two camels. 
On the other hand, the \netshort-Caption is able to identify the fact that there are two persons playing racquetball in Figure~\ref{fig:racquetball}.

\begin{figure}[t]
    \centering
    \small
    \begin{center}
        \includegraphics[width=1\linewidth]{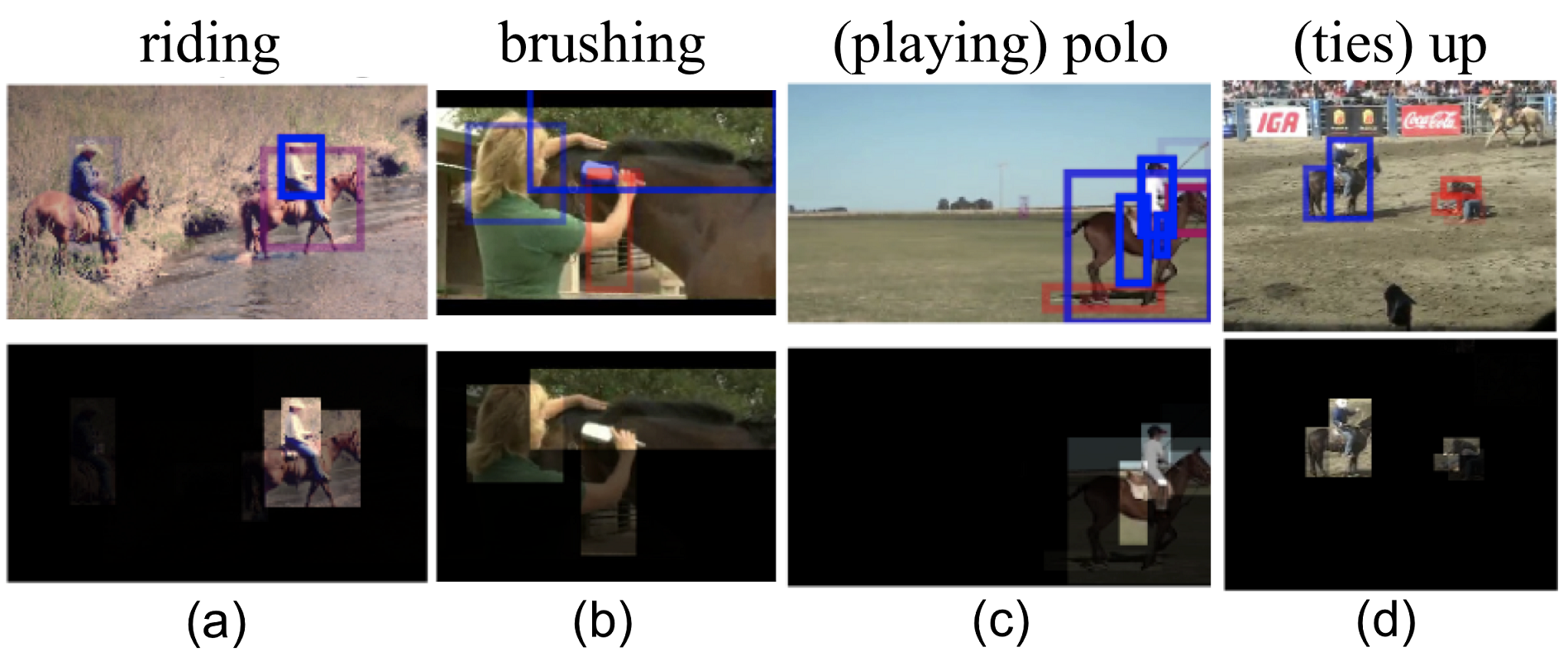}
    \end{center}
    \vspace{-0.1in}
    \caption{
    What interactions (verb) learned for video captioning.
    We verify how the \netshort-Caption distinguishes various type of interactions with a common object - \textit{horse}.
    (a) People are \textit{\underline{riding}} horses.
    (b) A woman is \textit{\underline{brushing}} a horse.
    (c) People are playing \textit{\underline{polo}} on a field.
    (d) The man \textit{\underline{ties}} up the calf.
    }
    \label{fig:verb}
    \vspace{-0.2in}
\end{figure}

\subsubsection{Distinguish interactions when common objects presented}
A common problem with the state-of-the-art captioning models is that they often lack the understanding of the relationships and interactions between objects, and this is oftentimes the result of dataset bias. For instance, when the model detects both person and a horse. The caption predictions are very likely to be: A man is riding on a horse, regardless whether if this person has different types of interactions with the horse.

We are thus interested in finding out whether if the proposed method has the ability to distinguish different types of interactions when common objects are presented in the scene. In Figure~\ref{fig:verb}, each video shares a common object in the scene - \textit{horse}. We show the verb (interaction) extracted from a complete sentence as captured by our proposed method. 

\begin{itemize}
    \item People are \textit{\underline{riding}} horses.
    \item A woman is \textit{\underline{brushing}} a horse.
    \item People are playing \textit{\underline{polo}} on a field.
    \item The man \textit{\underline{ties}} up the calf.
\end{itemize}

While all videos involve horses in the scene, our method successfully distinguishes the interactions of the human and the horse.

\subsubsection{Discussion on ActivityNet Captions}
We observed that while higher-order object interactions did contribute to higher performance on ActivityNet, the contributions were not as significant as when applied to the Kinetics dataset (quantitatively or qualitatively).
% We observed that the detected higher-order object interactions on ActivityNet Caption, though provides slightly higher performance, are quantitatively and qualitatively less influential compared with the results on Kinetics. 
We hereby discuss some potential reasons and challenges on applying \netshort-Caption on the ActivityNet Captions dataset.

\begin{figure*}[t]
    \centering
    \includegraphics[width=1\textwidth]{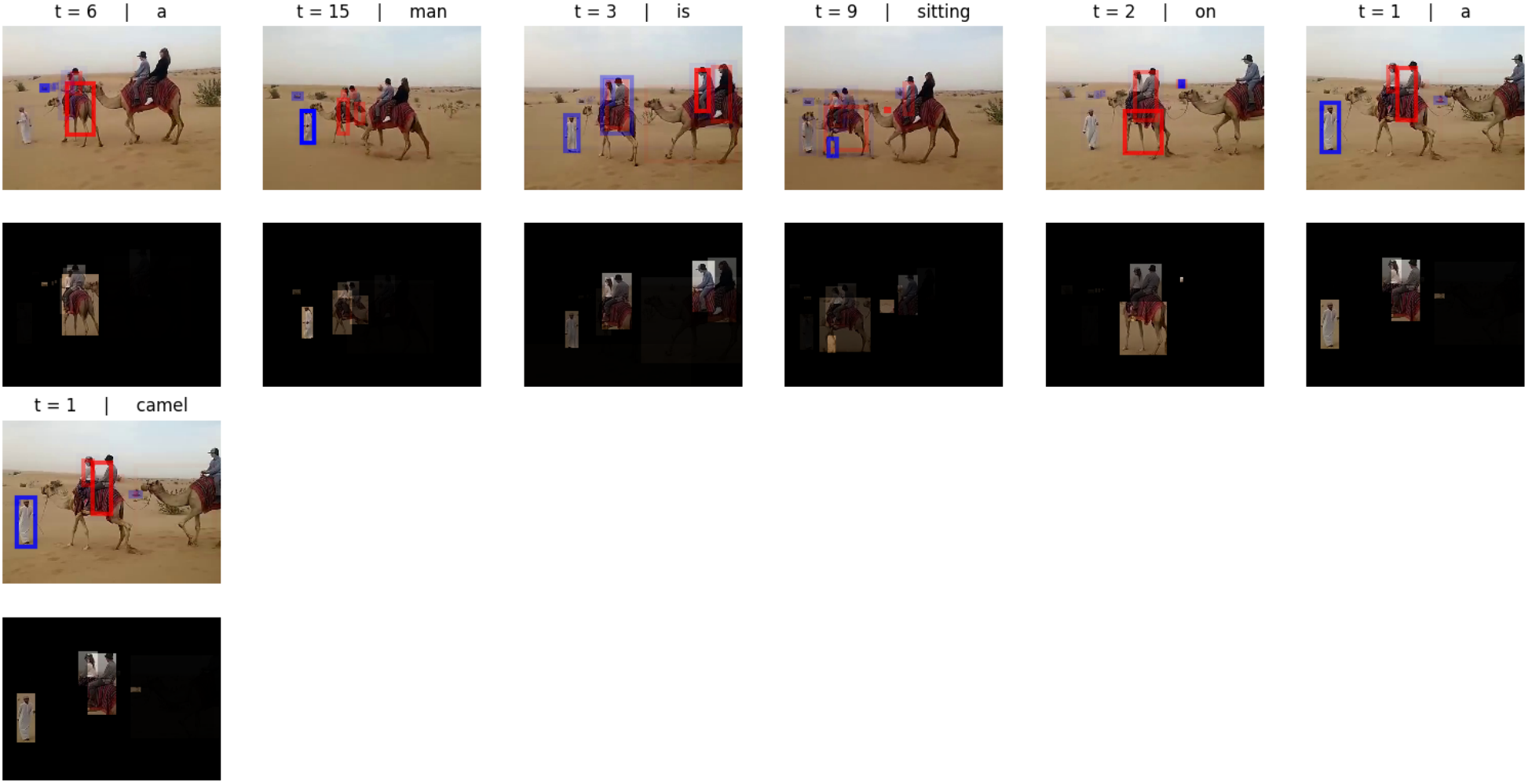}
    \caption{
    \textit{A man is sitting on a camel.}
    The \netshort-Caption is able to detect the ROIs containing both persons and the camel. 
    We can also observe that it highlights both the ROIs for persons who sit on the camel and the camel itself at frame 3 and 9. 
    However, the proposed method failed to identify that there are multiple people sitting on two camels. Furthermore, in some cases, it selects the person who leads the camels. 
    This seems to be because the same video is also annotated with another caption focusing on that particular person: \textit{A short person that is leading the camels turns around.} 
    }
    \label{fig:camel}
    \vspace{-0.1in}
\end{figure*}

\textbf{Word by word caption generation:}
In line with the work from question-answering, machine translation, and captioning, we generate a language sentence describing a video one word after another. 
At each word generation step, the \netshort-Caption uses the last generated word, video frame representations, and their corresponding object interactions.
As we can see from both qualitative results from Kinetics and ActivityNet Captions, our proposed method is able to identify the interactions within a very few video frames.
However, taking Figure~\ref{fig:camel} as an example, at the first word "a", our model has already successfully selected the persons (both in light blue and red) on top of the camel (bright red). 
Yet, during the following caption generation, the \netshort-Caption was \textit{forced} to look at the visual content again and again.
Introducing the gated mechanism~\cite{Lu2017Adaptive} may mitigate this issue, but our preliminary results do not show improvement.
Further experiments toward this direction may be needed.

\textbf{Semantically different captions exist:}
Each video in the ActivityNet Captions dataset consists of 3.65 (average) different temporal video segments and their own ground truth captions~\cite{krishna2017dense}. 
These video captions have different semantic meanings but oftentimes share very similar video content, i.e. the same/similar video content has several different ground truth annotations.
As a result, it may create confusion during the training of the model.
Again, taking Figure~\ref{fig:camel} as an example, we observed that the \netshort-Caption often focuses on the person who leads the camels ($t=1, 3, 15$).
We conjecture that this is due to the fact that, within the same video, there exists another video segment with annotation: \textit{A short person that is leading the camels turns around.}
Although within the same video content, one of the ground truth focuses on the persons sitting on the camels, another ground truth focuses on the person leading the camels.
This seems to be the reason why the trained network focuses on that particular person.
Based on this observation, we believe that future work in re-formulating these semantically different annotations of similar video content for network training is needed, and perhaps it may be a better way to fully take advantage of fine-grained object interactions detected from \netshort-Caption. 
One possibility will be associating semantically different video captions with different region-sequences within a video~\cite{shen2017weakly}. 

\subsection{Performance improvement analysis on Kinetics}
\begin{figure*}[t]
    \centering
    \small
    \begin{center}
        \includegraphics[width=1\linewidth]{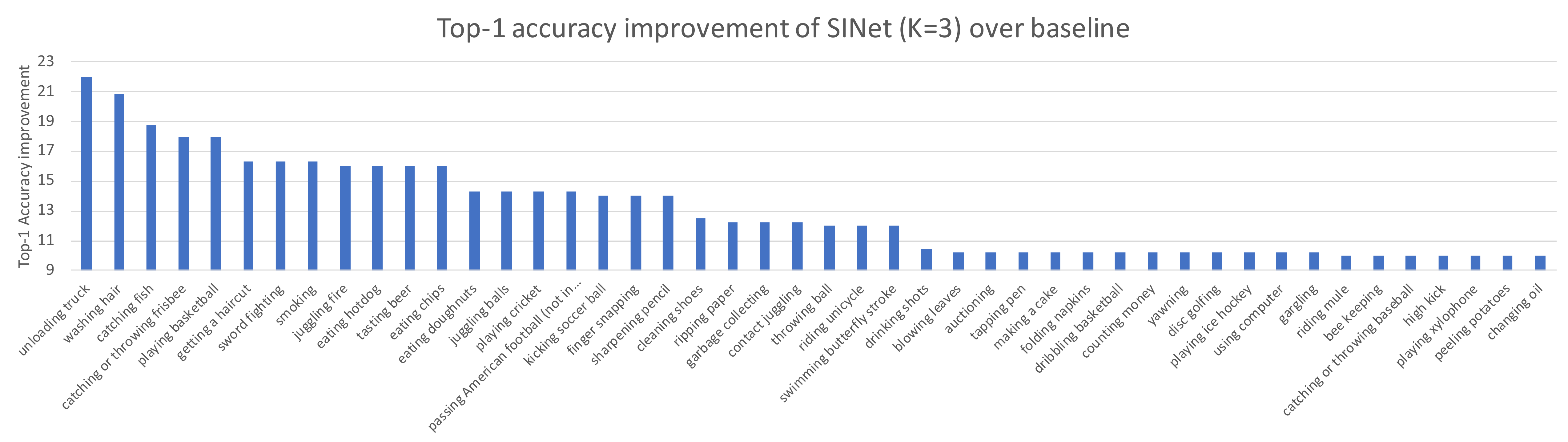}
    \end{center}
    \caption{
    Top-1 accuracy improvement of \netshort\ ($K=3$) over baseline. 46/400 classes that are improved more than 10\% are shown. 
    }
    \label{fig:improvement}
    % \vspace{-0.1in}
\end{figure*}

The proposed \netshort\ ($K=3$) shows more than 5\% improvement on top-1 accuracy in 136/400 classes and more than 10\% improvement in 46 classes over baseline. 
We show the classes that were improved more than 10\% on top-1 accuracy in Figure~\ref{fig:improvement}.
In addition to these classes, the proposed \netshort\ in modeling fine-grained interactions specifically improved many closely related classes. 

\begin{itemize}
  \item 7 classes related to \textbf{hair} that are ambiguous among each other: \textit{braiding hair}, \textit{brushing hair}, \textit{curling hair}, \textit{dying hair}, \textit{fixing hair}, \textit{getting a haircut}, and \textit{washing hair}. We show 21\% top-1 improvement on \textit{washing hair}; 16\% improvement on \textit{getting a haircut}.
  \item 4 classes related to \textbf{basketball} require the model to identify how the basketball are being interacted. These classes are: \textit{playing basketball}, \textit{dribbling basketball}, \textit{dunking basketball}, and \textit{shooting basketball}. We observed 18\%, 10\%, 6\%, and 8\% improvement respectively. 
  \item Among 3 related to \textbf{juggling} actions: \textit{juggling fire}, \textit{juggling balls}, and \textit{contact juggling}. We obtained 16\%, 14\%, and 13\% improvement respectively. 
  \item Our model significantly improved the \textbf{eating} classes, which are considered to be the hardest~\cite{kay2017kinetics}, because they require distinguishing what is being eaten (interacted). We show improvement among all eating classes, including \textit{eating hot dog}, \textit{eating chips}, \textit{eating doughnuts}, \textit{eating carrots}, \textit{eating watermelon}, and \textit{eating cake}. 
  We obtained 16\%, 16\%, 14\%, 8\%, 4\%, and 4\% improvement respectively.
%   \item Note that “eating cake” can also be confused with “making a cake”, since both involving cakes in the scene. By successfully identify the interaction of objects, we further obtained 10\% improvement on identifying “making a cake” class. 
\end{itemize}

% For example (all \% are absolute improvement averaged over similar classes), all 4 classes that require the model to  identify how a basketball being interacted with were improved 10\%, such as playing basketball, dribbling basketball, dunking basketball, and shooting basketball. 
% Similar improvements are also seen in other classes: hair (18\%), juggling (13\%) and eating classes (10\%). 

\subsection{ActivityNet Captions on 1st and 2nd val set}
\label{sec:caption-vals-results}
We report the performance of \netshort-Caption\ on the 1st and the 2nd validation set in Table~\ref{table:caption-two-eval}.
We can see that using fine-grained (higher-order) object interactions for caption generation consistently shows better performance than using coarse-grained image representation, though the difference is relatively minor compared to the results on Kinetics.
We discuss the potential reasons in Sec.~\ref{sec:discussion-Activitynet}.
Combining both coarse- and fine-grained improve the performance across all evaluation metrics. 
Interestingly, using co-attention on detected object interactions shows better performance on the 1st validation set but has similar performance on the 2nd validation set. 

\subsection{Model architecture and FLOP}
\label{sec:sub-FLOPs}
We now describe the model architecture of the proposed recurrent higher-order module and how the FLOP is calculated. 

\textbf{\netshort\ architecture:}
We first project the image representations $v_{c,t}$ to introduce learnable feature representations. The MLP $g_\phi$ consist of two sets of fully-connected layers each with batch normalization and ReLU. It maintains same dimension ($m=2048$) of the input image feature.
Thus, the coarse-grained representation of the video is a feature vector with 2048 dimension.
Inside the Recurrent HOI module, each of the MLP $g_{\theta_k}$ has three sets of batch normalization layers, fully-connected layers, and ReLUs. 
In the experiments with two attentive selection module ($K=2$), we set the dimension of the fully-connected layer to be 2048. 
The concatenation of $v_{{o, t}}^1$ and $v_{{o, t}}^2$ is then used as the input to the following LSTM cell. 
Empirically, we find out that it's important to maintain high dimensionality for the input to LSTM cell.
We adjust the dimension of hidden layers in $g_{\theta_k}$ given the number of $K$, e.g. we reduce the dimension of the hidden layer if $K$ increases.
In this way, the inputs to LSTM cell have the same or similar feature dimension for fair experimental comparison. 
The hidden dimension of the LSTM cell is set to be 2048. 
Before concatenating the coarse- ($v_c$) and fine-grained ($v_{oi, T}$) video representations, we re-normalize the feature vector with batch normalization layer separately. 
The final classifier then projects the concatenated feature representation to 400 action classes.

% \begin{figure*}[t]
%     \centering
%     \includegraphics[width=1\textwidth]{figure/qualitative/jetskiing.pdf}
%     \caption{
%     \textbf{Jetskiing}. At the beginning, human subject is highlighted. The attention then shifts to the relationships between human, jet ski, and the sea in the background, and successfully identify \textit{Jetskiing}.
%     }
%     \label{fig:jetskiing}
% \end{figure*}

% \begin{figure*}[t]
%     \centering
%     \includegraphics[width=1\textwidth]{figure/qualitative/testifying.pdf}
%     \caption{
%     \textbf{Testifying} is ambiguous since there are many other action classes involving a human talking straight to the camera. For instance, \textit{Auctioning}, \textit{Sign language interpreting}, \textit{Presenting weather forecast}, \textit{News anchoring}, \textit{Answering questions}, etc. 
%     It is interesting to see that our \netshort\ not only utilize the person at the center and the crowd in the background, but also heavily rely on the desk name tag and the water bottle as visual cues for predicting \textit{Testifying}. 
%     }
%     \label{fig:testifying}
% \end{figure*}

% \begin{figure*}[t]
%     \centering
%     \includegraphics[width=1\textwidth]{figure/qualitative/water_surfing_1.pdf}
%     \caption{
%     \textit{People are surfing on the water.}
%     At the beginning, the \netshort\ identify multiple people are in the ocean. The person who is surfing on the water is then successfully identified, and the rest of irrelevant objects and background are completely ignored.
%     }
%     \label{fig:water_surfing_1}
%     \vspace{-0.1in}
% \end{figure*}

\begin{table*}[t]
    \centering
    \caption{
        METEOR, ROUGE-L, CIDEr-D, and BLEU@N scores on the ActivityNet Captions 1st and 2nd validation set. All methods use ground truth temporal proposal, and out results are evaluated using the code provided in \cite{krishna2017dense} with $tIoU=0.9$.
        Our results with ResNeXt spatial features use videos sampled at maximum 1 FPS only.
        }
    \label{table:caption-two-eval}
    \small
    \begin{tabular}{lccccccc}
    Method  & B@1   & B@2  & B@3 & B@4    & ROUGE-L    & METEOR    & CIDEr-D   \\ \hline
    \multicolumn{1}{c}{\textbf{1st Validation set}} & & & & & & & \\ \hline
    \netshort-Caption | img (C3D)    & 16.93 & 7.91 & 3.53  & 1.58  & 18.81 & 8.46  & 36.37 \\
    \netshort-Caption | img (ResNeXt)    & 18.71 & 9.21  & 4.25  & 2.00  & 20.42 & 9.55  & 41.18 \\
    \netshort-Caption | obj (ResNeXt)    & 19.00 & 9.42  & 4.29  & 2.03  & 20.61 & 9.50  & 42.20 \\
    \netshort-Caption | img + obj | no co-attention (ResNeXt)    & \textbf{19.89} & 9.76  & 4.48  & 2.15  & 21.00 & 9.62  & 43.24 \\
    \netshort-Caption | img + obj (ResNeXt)     & 19.63    & \textbf{9.87}   & \textbf{4.52}   & \textbf{2.17} & \textbf{21.22}    & \textbf{9.73} & \textbf{44.14} \\ \hline
    \multicolumn{1}{c}{\textbf{2nd Validation set}} & & & & & & & \\ \hline
    \netshort-Caption | img (C3D)    & 17.42 & 8.07 & 3.53  & 1.35  & 18.75 & 8.41  & 40.06 \\
    \netshort-Caption | img (ResNeXt)    & 18.91 & 9.41  & 4.28  & 1.68  & 20.49 & 9.56  & 45.05 \\
    \netshort-Caption | obj (ResNeXt)    & 19.14 & 9.53  & 4.47  & 1.81  & 20.73 & 9.61  & 45.84 \\
    \netshort-Caption | img + obj | no co-attention (ResNeXt)    & \textbf{19.97} & 9.88  & \textbf{4.55}  & \textbf{1.90}  & 21.15 & \textbf{9.96}  & \textbf{46.37} \\
    \netshort-Caption | img + obj (ResNeXt)     & 19.92    & \textbf{9.90}   & 4.52   & 1.79 & \textbf{21.28}    & 9.95 & 45.54
\end{tabular}
\end{table*}

\textbf{\netshort-Caption architecture:}
We first use a single fully-connected layer with batch normalization, dropout, and ReLU to project the pre-saved image features $v_{c,t}$. The $g_\phi$ maps the feature vector from 2048 to 1024. 
We use two attentive selection modules for video captioning task ($K=2$).
Each $g_{\theta_k}$ consist of a batch normalization, fully-connected layer, dropout layer, and a ReLU. It maps input object feature vector from 2048 to 512. 
The dropout ratio for both $g_\phi$ and $g_{\theta_k}$ are set to be 0.5.
The concatenation of $v_{{o, t}}^1$ and $v_{{o, t}}^2$ is used as input to the LSTM cell inside Recurrent HOI module. 
The hidden dimension of this LSTM cell is set to be 1024. 
The dimension of word embedding is 512. We use ReLU and dropout layer after embedding layer with dropout ratio 0.25. 
The hidden dimension of both Attention LSTM and Language LSTM are set to be 512. 

\textbf{FLOP}
is computed per video and the maximum number of objects per frame is set to 15. 
We compare the computed FLOP with traditional object interactions by paring all possible objects. 
The results are shown in Table~\ref{tabel:FLOPs}.

\begin{table*}[!t]
    \centering
    \caption{FLOPs calculation on Kinetics sampled at 1 FPS. The calculation is based on forward passing of one video.}
    \label{tabel:FLOPs}
    \begin{tabular}{c|c|c|c|c|c}
    \multicolumn{2}{c|}{Proposed method (\textcolor{blue}{$K=2$})}                                                                   & FLOP & \multicolumn{2}{c|}{Object pairs} & FLOP  \\ \hline
    \multicolumn{6}{c}{Project obj features}                                                                                                                     \\ \hline
    \multirow{3}{*}{MLP $g_{\theta_k (o_{i,t})}$}         & 15 x 2048 x 2048 x \textcolor{blue}{2}                                                                & 0.13e9  & \multirow{3}{*}{MLP}    & 105 x 4096 x 2048      & 0.9e9  \\
    & 15 x 2048 x 2048 x \textcolor{blue}{2}                                                                & 0.13e9   &    & 105 x 2048 x 2048       & 0.4e9  \\ 
    & 15 x 2048 x 2048 x \textcolor{blue}{2}                                                                & 0.13e9   &    & 105 x 2048 x 2048      & 0.4e9  \\ \hline
    \multicolumn{6}{c}{Recurrent unit}                                                                                                                                     \\ \hline
    \multicolumn{2}{c}{Recurrent HOI (SDP-Attention)}                                               &        &            &                        &        \\ \hline
    $W_h h_{t-1}$    & 2048 x 2048 x \textcolor{blue}{2}                                                                         & 8.4e6   &            &                        &        \\
    $W_c v_{c,t}$ & 2048 x 2048 x \textcolor{blue}{2}                                                                         & 8.4e6   &            &                        &        \\
    MatMul          & 15 x 15 x 2048 x \textcolor{blue}{2}                                                                  & 0.9e6 &            &                        &        \\
    MatMul          & 15 x 15 x 2048 x \textcolor{blue}{2}                                                                  & 0.9e6 &            &                        &        \\ \hline
    LSTM Cell            & 8 x 2 x \textcolor{blue}{2} x 2048 x 2048 & 134.2e6 & LSTM Cell           & 8 x 2 x 2048 x 2048    & 67e6   \\ \hline
    \multicolumn{6}{c}{Total}                                                                                                                                    \\ \hline
    timesteps (\textcolor{green}{$T=10$})  & \textcolor{green}{10} x (MLP + Recurrent)                                                                         & 5.3e9   &            & \textcolor{green}{10} x (MLP + Recurrent)               & 18.3e9
    \end{tabular}
\end{table*}

\clearpage{\thispagestyle{empty}\cleardoublepage}
{\small
\bibliographystyle{ieee}
\bibliography{egbib}
}

\end{document}